\documentclass[10pt,journal,compsoc]{IEEEtran}

\ifCLASSOPTIONcompsoc
  \usepackage[nocompress]{cite}
\else
  \usepackage{cite}
\fi

\ifCLASSINFOpdf
\else
\fi

\usepackage{microtype}
\usepackage{graphicx}
\usepackage{booktabs} % for professional tables
\usepackage{hyperref}

\usepackage{subcaption}
\usepackage[normalem]{ulem}
\usepackage{microtype}
\usepackage{graphicx}

\usepackage{url}

\usepackage{breakurl}

\usepackage{tabu}
\usepackage{booktabs} % for professional tables
\usepackage{amsmath}
\usepackage{amsfonts}
\usepackage{eucal}
\usepackage{xspace}
\usepackage{bm}
\usepackage{enumitem}
\usepackage{color}
\usepackage{xcolor}
\usepackage{soul}
\usepackage[ruled,linesnumbered]{algorithm2e}
\SetKwComment{Comment}{$\triangleright$\ }{}
\usepackage[flushleft]{threeparttable}
\usepackage{stfloats} % position two-col figs
\usepackage{nicefrac}
\usepackage{mathtools}
\usepackage{array,multirow}
\usepackage{tikz}
\usepackage{pifont}
\newcommand{\cmark}{\ding{51}}%
\newcommand{\xmark}{\ding{55}}%

\newcommand{\method}{NAWQ-SR\xspace}
\newcommand{\tool}{NAWQ-SR\xspace}

\newlength\mylenin
\newcommand\myinput[1]{%
\settowidth\mylenin{\KwIn{}}%
\setlength\hangindent{\mylenin}%
\hspace*{\mylenin}#1\\}

\usepackage{textgreek} % for uLayer reference from EuroSys 2019.

\let\oldnl\nl % Store \nl in \oldnl
\newcommand{\nonl}{\renewcommand{\nl}{\let\nl\oldnl}} % Remove line number for one line

\newlength\mylenout
\newcommand\myoutput[1]{%
\settowidth\mylenout{\KwOut{}}%
\setlength\hangindent{\mylenout}%
\hspace*{\mylenout}#1\\}

\makeatletter
\def\footnoterule{\relax%
  \kern-5pt
  \hbox to \columnwidth{\hfill\vrule width 1\columnwidth height 0.4pt\hfill}
  \kern4.6pt}
\makeatother

\newcommand{\blue}[1]{\textcolor{black}{#1}}

\newcommand{\red}[1]{\textcolor{black}{#1}}

\usepackage{background}

\backgroundsetup{angle=0, 
    scale=1,
    color=black,
    firstpage=true,
    position=current page.north,
    hshift=0pt,
    vshift=-20pt,
    contents={\ifnum\value{page}=1 PREPRINT: Accepted for publication at the IEEE Transactions on Mobile Computing (TMC), 2023 \else \fi}
}

\begin{document}
\title{\method: A Hybrid-Precision NPU Engine for Efficient On-Device Super-Resolution}

\author{Stylianos~I.~Venieris, %~\IEEEmembership{Member,~IEEE,}
        Mario~Almeida,
        Royson~Lee,
        and~Nicholas~D.~Lane% <-this % stops a space
% \IEEEcompsocitemizethanks{\IEEEcompsocthanksitem M. Shell was with the Department
% of Electrical and Computer Engineering, Georgia Institute of Technology, Atlanta,
% GA, 30332.\protect\\
% % note need leading \protect in front of \\ to get a newline within \thanks as
% % \\ is fragile and will error, could use \hfil\break instead.
% E-mail: see http://www.michaelshell.org/contact.html
% \IEEEcompsocthanksitem J. Doe and J. Doe are with Anonymous University.}% <-this % stops an unwanted space
% \thanks{Manuscript received April 19, 2005; revised August 26, 2015.}
\IEEEcompsocitemizethanks{\IEEEcompsocthanksitem Stylianos I. Venieris is with the Samsung AI Center, CB1 2JH Cambridge, U.K.
\IEEEcompsocthanksitem Mario Almedia is with Rain Instant Pay.
\IEEEcompsocthanksitem Royson Lee and Nicholas D. Lane are with the University of Cambridge, CB2 1TN Cambridge, U.K. and also with the Samsung AI Center, CB1 2JH Cambridge, U.K.%}% <-this % stops a space
}
% \thanks{Preprint: Under review.}
}

\IEEEtitleabstractindextext{%
\begin{abstract}
    In recent years, image and video delivery systems have begun integrating deep learning super-resolution (SR) approaches, leveraging their unprecedented visual enhancement capabilities while reducing reliance on networking conditions.
    Nevertheless, deploying these solutions on mobile devices still remains an active challenge as SR models are excessively demanding with respect to workload and memory footprint.
    Despite recent progress on on-device SR frameworks, existing systems either penalize visual quality, lead to excessive energy consumption or make inefficient use of the available resources. 
    This work presents \tool, a novel framework for the efficient on-device execution of SR models.
    Through a novel hybrid-precision quantization technique and a runtime neural image codec, \tool exploits the multi-precision capabilities of modern mobile NPUs in order to minimize latency, while meeting user-specified quality constraints. 
    Moreover, \tool selectively adapts the arithmetic precision at run time to equip the SR DNN's layers with wider representational power, improving visual quality beyond what was previously possible on NPUs.
    \blue{Altogether, \tool achieves an average speedup of 7.9$\times$, 3$\times$ and 1.91$\times$ over the state-of-the-art on-device SR systems that use heterogeneous processors (MobiSR), CPU (SplitSR) and NPU (XLSR), respectively.}
    \blue{Furthermore, \tool delivers an average of 3.2$\times$ speedup and 0.39 dB higher PSNR over status-quo INT8 NPU designs, but most importantly mitigates the negative effects of quantization on visual quality, setting a new state-of-the-art in the attainable quality of NPU-based SR.} 
\end{abstract}

}

\maketitle

\IEEEdisplaynontitleabstractindextext

% \fi
\IEEEpeerreviewmaketitle

\vspace{-0.2cm}
\IEEEraisesectionheading{\section{Introduction}
\label{sec:intro}}
% \vspace{-0.025cm}

With the rapid rise of Internet content delivery services and devices that support higher resolution content, images and videos are predicted to account for 82\% of the global Web traffic~\cite{ciscovni}. 
Mobile applications, in particular, constitute a great proportion of this growth, as services such as live streaming, video-conferencing, and video-on-demand have been on the rise. 
For instance, popular video app {\small TikTok} has over 50 million daily users 
with increases of 55\% in unique users and 93.7\% in the average time spent per user in just six months~\cite{tiktok_stats}. 
To meet such demands, mobile systems are required to maximize both the user satisfaction and their quality of experience (QoE).

A primary challenge of this class of mobile systems is their sensitivity to networking conditions. 
In real-world cellular networks, the network speed fluctuates substantially, and poor connectivity leads to excessive response times, dropped frames or video stalling, which rapidly degrade the QoE~\cite{videoqoe2011inm,videoqoe2009itng,videoqoe2016imc,Yin2015sigcomm}. This phenomenon is further aggravated by the increasing number of users which compete for the same pool of network resources and create contention~\cite{netpolice2009imc}.

A recent key method to handle the aforementioned drawbacks is \textit{neural enhancement} via super-resolution (SR) deep neural networks (DNNs)~\cite{SRCNN}.
SR DNNs operate by processing a low-resolution, degraded image to automatically generate a high-quality, high-resolution output. 
This allows compact, low-quality content to be transmitted across the network, at the expense of additional computation at the receiver's end.
As such, neural enhancement removes the system's sole reliance on the network and opens up a new dimension in the design space by introducing a trade-off between the use of bandwidth and computational resources~\cite{neuralenhancement2020,dnn_cds2021csur}.

Despite the increasing processing capabilities of mobile devices, \textit{on-device} execution of SR models still remains an active challenge due to their demanding workload.
In particular, the number of multiply-add operations and memory capacity required even by mobile-tailored SR DNNs is \textit{\mbox{orders} of magnitude larger} than the more common classification DNNs~\cite{mobisr2019mobicom}.
To counteract the excessive computational needs, existing systems \textit{1)}~rely on powerful platforms,
% floating-point implementations, 
such as assuming the availability of a desktop GPU client~\cite{video_delivery2018osdi, dejavu2019hotmobile}, \textit{2)}~require the parallel use of all available processors (CPU, GPU, NPU)~\cite{mobisr2019mobicom}, \textit{3)}~leverage frame dependencies in order to cache previously enhanced results~\cite{nemo2020mobicom} or \textit{4)}~resort to cloud offloading~\cite{supremo2020tmc}.
As such, existing solutions are either restricted to high-end deployment settings~\cite{video_delivery2018osdi, dejavu2019hotmobile}, thus not accommodating mobile devices, or incur additional issues as a by-product, such as thermal throttling~\cite{mitra2019tcad,fire_in_your_hand2019mobicom,mobisr2019mobicom} and a drastic drop in visual quality~\cite{nemo2020mobicom, supremo2020tmc}. 

\begin{figure}[t]
    % \vspace{0.3cm}
    \captionsetup[subfigure]{labelformat=empty}
    \centering
    \subfloat[]{
        \includegraphics[
        width=0.49\textwidth]{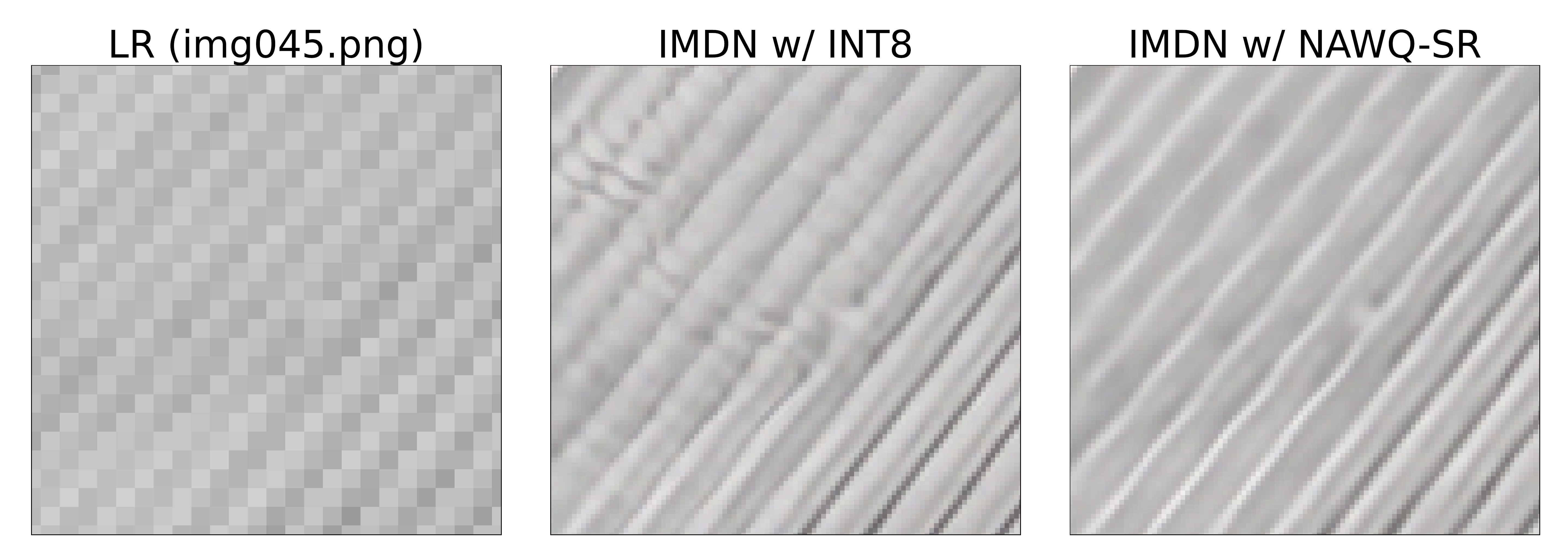}
    }\\
    \vspace{-0.6cm}
    \centering
    \subfloat[]{
        %\label{main:c}
        \includegraphics[
        %trim={1cm 2cm 0.5cm 1.5cm},clip,
        width=0.49\textwidth]{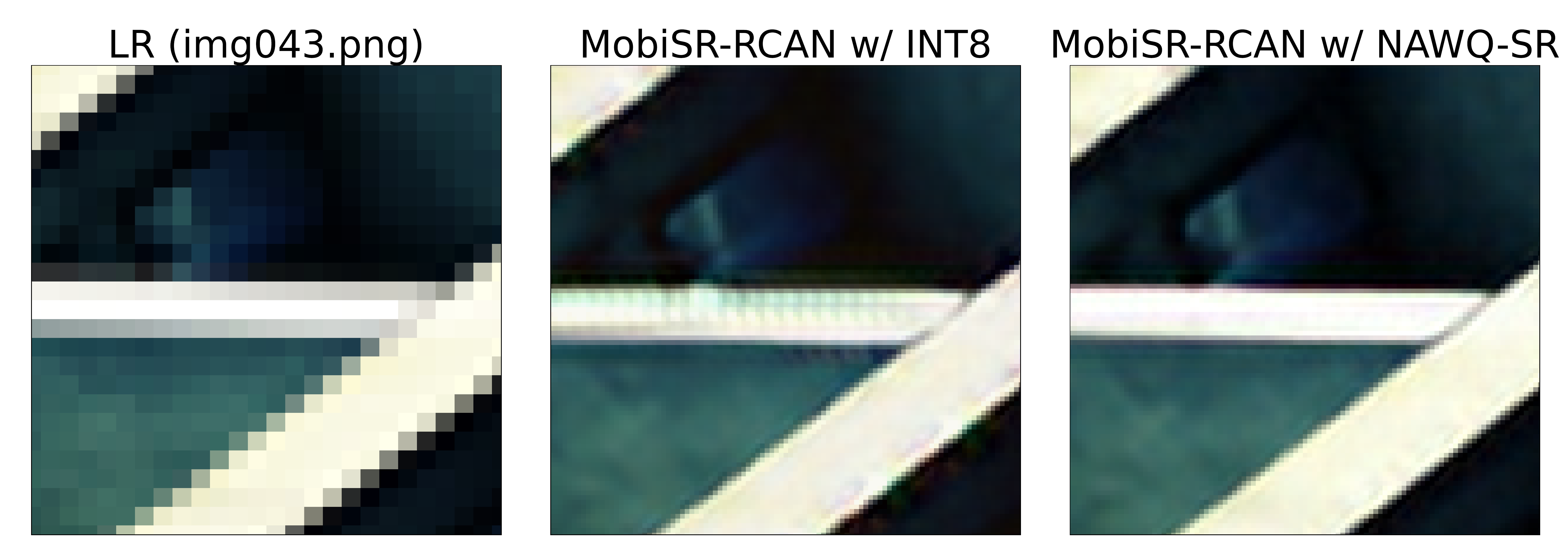}
    }\\
    % \vspace{-1cm}
    \vspace{-0.6cm}
    \centering
    \subfloat[]{
        \includegraphics[
        width=0.49\textwidth]{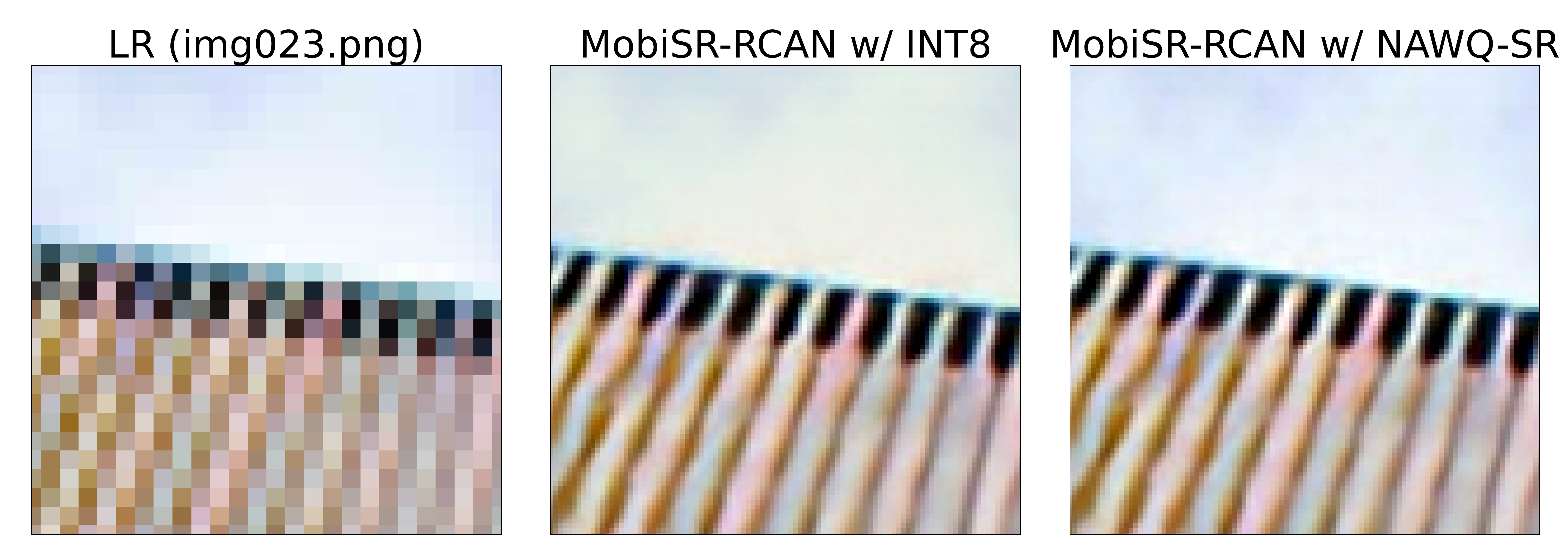}
    }\\
    \vspace{-0.6cm}
    \centering
    \subfloat[]{
        \includegraphics[
        width=0.49\textwidth]{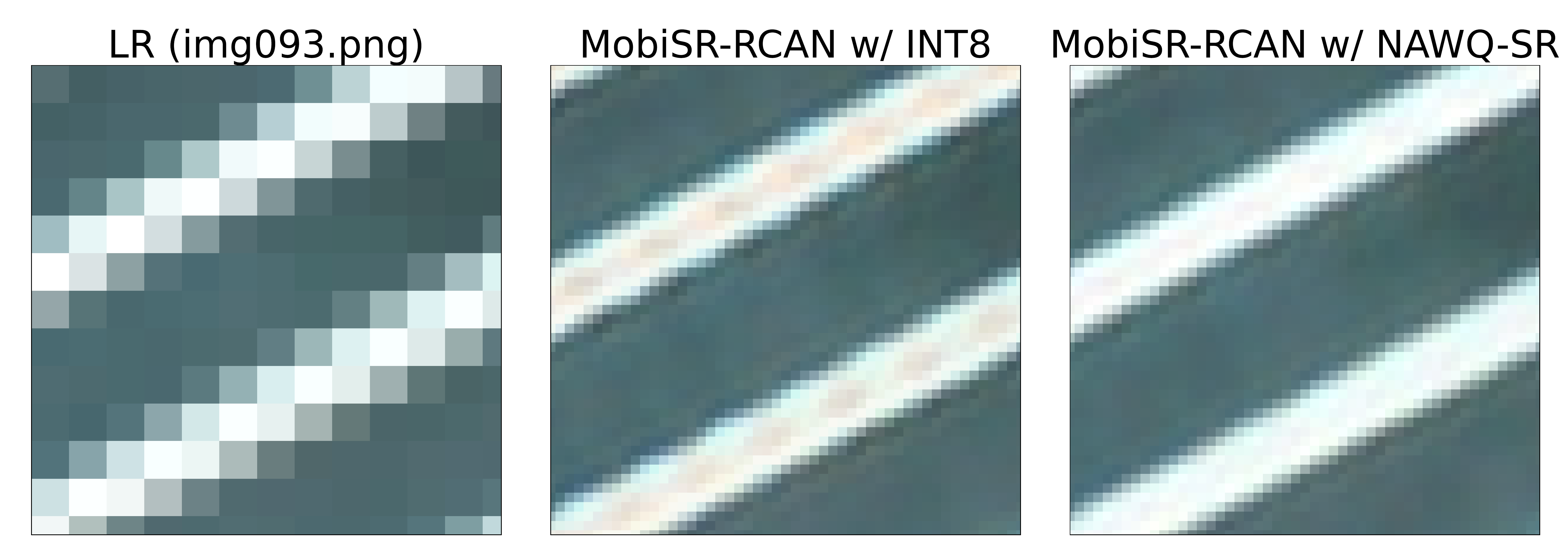}
    }\\
    \vspace{-0.7cm}
    \caption{Qualitative comparison between INT8 and \tool $\times4$ upscaling on the Urban100~\cite{Urban100} dataset. INT8 results in visual artifacts on both textures and colors when compared to \tool's hybrid-precision and DRE-based quality enhancement.
    }
    \vspace{-0.4cm}
    \label{fig:qualitative}
\end{figure}

To counteract these limitations and enable the use of SR DNNs on mobile devices, there has been an increased focus towards low-precision DNN execution on faster and more efficient processing units like NPUs~\cite{mobisr2019mobicom,xlsr2021cvprw}.
These units provide higher energy efficiency than CPUs and GPUs by omitting general-purpose hardware logic, increasing at the same time the availability of computational resources for other tasks by taking over the compute-intensive DNN execution. 
Despite the NPUs' demonstrated benefits for \textit{classification} DNNs, executing SR models at lower precision often comes at the cost of degraded visual quality; 
\blue{as shown in Fig.~\ref{fig:qualitative}, upscaling with INT8 - \textit{which is conventionally thought to be the best data type for inference} - would result in unnatural visual artifacts on both the texture and color in some images, especially for deeper DNNs.}
%\blue{ despite marginal quantitative loss in standard metrics (\textit{e.g.} PSNR/SSIM) as seen in Fig.~\ref{fig:qualitative}. }
\blue{Notably, these anomalies could also happen despite marginal quantitative loss in standard metrics (0.1~dB drop in PSNR) as minor differences at the pixel level can still result in considerable high-level deformation.}
\blue{As a result, existing on-device SR frameworks such as MobiSR~\cite{mobisr2019mobicom} and NEMO~\cite{nemo2020mobicom} underutilize or entirely avoid execution on the NPU to meet an acceptable visual quality.}
% As discussed in Section~\ref{sec:background}, this is especially manifested in state-of-the-art SR DNNs, where quantization can lead to an unacceptable drop in visual quality~\cite{mobisr2019mobicom,nemo2020mobicom,PAMSQS}.
Thus, there is an emerging need for novel solutions that allow leveraging mobile NPUs for SR without the quality impact of low-precision data types. 

In this work, we present \tool, a framework that overcomes the limitations of existing on-device SR systems and delivers fast, efficient and high-quality SR on mobile.
% To optimize latency while meeting the user-specified quality constraints,
\tool introduces an NPU-centric approach, comprising a novel \textit{hybrid-precision execution} paradigm and a runtime \textit{neural image codec} that exploit the multi-precision processing capabilities of modern mobile NPUs to minimize latency while meeting the user-specified quality targets. 
Moreover, to push visual quality beyond the state-of-the-art NPU-based designs, we propose a mechanism that selectively re-customizes the arithmetic precision of the DNN layers on-the-fly.
% , improving visual quality beyond the state-of-the-art NPU-based designs.
This paper makes the following key contributions:
\vspace{-0.1cm}
\begin{itemize}[leftmargin=*]
    \setlength\itemsep{-0.5mm}
    \item A novel hybrid-precision execution scheme together with a methodology for optimizing the deployment of SR DNNs to the latency and quality requirements of the target application.
    By considering the multiple precisions supported by a given NPU, our framework adapts each layer's word\-length through a single-shot optimization algorithm that co-optimizes the per-layer quantization of the DNN and the scheduling of its layers on the NPU. 
    
    \item A novel technique that selectively applies adaptive arithmetic precision on quantization-sensitive layers, enhancing them with wider representational power at run time. We dynamically adapt the quantization parameters of the selected layers in a per-sample input-dependent manner, leading to lower quantization error and higher visual quality than previously attainable on mobile NPUs. 
    % \item A novel technique that identifies quantization-sensitive layers and selectively applies adaptive arithmetic precision, enhancing them with wider representational power at run time. We dynamically adapt the quantization parameters of the selected layers in an input-dependent manner, leading to lower quantization error and higher visual quality than previously attainable on mobile NPUs. 
    
    \item \blue{A new neural image codec comprising a hybrid-precision dispatcher and a runtime quantization unit. 
    % The codec is configured with the SR DNN-optimized hybrid-precision configuration and the associated execution schedule, and provides optimized execution on mobile. 
    Through our low-overhead codec, we provide a \textit{fully NPU-based} execution of SR DNNs that avoids barriers of current NPU support for upsampling layers, acknowledged by previous works, that conventionally required CPU or GPU fallback.}
    % Additionally, our design overcomes existing obstacles, acknowledged by previous works, of current NPU support, such as the NPU-based mapping of upscaling layers. }
    
    \item To the best of our knowledge, this work is the first SR approach to exploit the multi-precision capabilities of the heterogeneous processing units that reside in NPUs.  
    \blue{Hence, it can be orthogonally combined with existing on-device SR systems such as MobiSR~\cite{mobisr2019mobicom} to counteract their performance limitations on the NPU.}
    \blue{As a standalone framework, it delivers} a speedup of 1.6$\times$-9.8$\times$ over state-of-the-art on-device SR systems and 91\% over XLSR, the winner of the Mobile AI 2021 challenge on real-time quantized SR.
\end{itemize}

\vspace{-0.25cm}
\section{Background \& Related Work}
\label{sec:background}
% \vspace{-0.1cm}

In this section, we discuss the emerging use of super-resolution for efficient visual enhancement on mobile devices, model- and system-level optimizations for the on-device execution of SR models and the main characteristics of the latest mobile NPUs.

\subsection{Super-resolution for Mobile Devices}

The unprecedented performance of SR DNNs in restoring realistic textures, together with their orthogonal integration with image/video compression and adaptive bitrate schemes, has made them a key component behind a broad range of products, from high-resolution TVs~\cite{samsung_tv_ai_upscaling} to gaming GPUs~\cite{nvidia_dynamic_sr}.
As such, several works have focused on 
improving the quality of mapping low-resolution (LR) images to high resolution (HR)~\cite{RCAN, IDN, rethinkdataaugyoocvpr2020}.
Despite the significant progress~\cite{USRNet, DAN}, SR DNNs still have prohibitively high computational and memory demands for most real-world mobile deployments.

\textbf{Efficient Super-resolution.}
Recent works have proposed efficiency-optimized model architectures.
Prominent %hand-crafted 
techniques span from avoiding the computation of large feature maps~\cite{SRCNN} and mitigating the cost of upsampling through the use of pixel-shuffle layers~\cite{ESPCN,FEQE}, to employing more efficient blocks, such as group convolutions~\cite{CARN} and channel splitting~\cite{IDN,IMDN}. 
% Apart from manual efforts, 
Neural architecture search for efficient SR is also gaining traction~\cite{tpsr2020eccv, FALSR, ESRN}.
% NAS methods span from multi-objective search schemes~\cite{MoreMNAS, FALSR} that co-optimize image fidelity, compute and memory, to minimizing search time~\cite{ESRN} by considering coarse building blocks, instead of more primitive operations.
% More recently, Lee \textit{et al.}~\cite{tpsr2020eccv} proposed a generative adversarial network (GAN) search and found a tiny model, named TPSR, that can maximize either image fidelity or perceptual quality.
Nonetheless, the on-device execution of these models is still impractical, resulting in numerous system-based solutions~\cite{dnn_cds2021csur}.

\textbf{On-device Super-resolution.}
% Running SR on mobile phones can drastically reduce the bandwidth requirements and the corresponding cellular data cost for both the users and the service provider across diverse applications. Such apps span from video-on-demand~\cite{video_delivery2018osdi} and video-conferencing~\cite{dejavu2019hotmobile} to graphics enhancement in mobile game streaming~\cite{stadia2021}.
% 
To deploy SR models on mobile, the state-of-the-art on-device SR frameworks have adopted various approaches. 
One line of work~\cite{mobisr2019mobicom,supremo2020tmc} has focused on utilizing the heterogeneous processors (CPU, GPU, NPU) residing in many recent devices. 
To effectively load-balance across processors, these systems exploit the observation that patches of an image have varying upsampling difficulty. 
For instance, MobiSR~\cite{mobisr2019mobicom} adopts a criterion to quantify the difficulty of each patch and dispatch it to the appropriate processor.
Besides scheduling, the video-focused NEMO~\cite{nemo2020mobicom} leverages the inter-frame dependencies in order to cache and reuse previously super-resolved patches.
Finally, SplitSR~\cite{splitsr2021imwut} combined efficient model design with compiler optimizations to improve CPU-based SR and XLSR~\cite{xlsr2021cvprw} presented a hand-crafted lightweight model.

Even though these frameworks enable fast on-device upsampling, they come at the high cost of \textit{quality degradation}. 
Notably, mapping these models on compute engines that run on lower bitwidths, such as NPUs, causes a considerable drop in visual quality as observed in recent mobile SR systems~\cite{mobisr2019mobicom,nemo2020mobicom,xlsr2021cvprw}.
As a result, existing systems either reduce the number of patches dispatched to NPUs~\cite{mobisr2019mobicom} or entirely avoid using them~\cite{nemo2020mobicom,splitsr2021imwut}, leading to reduced efficiency compared to NPU-only execution. 
As little work has been done to mitigate the effects of quantization on SR models, our work aims to breach this gap to allow existing 
techniques to leverage the full capabilities of modern NPUs that can be found across smartphones~\cite{embench2019emdl,almeida2021smart,ai_benchmark2019iccvw,snpe}.

\textbf{Quantization.}
Precision quantization constitutes a prominent method for minimizing the computational and memory demands of DNNs.
State-of-the-art approaches typically adopt block floating-point schemes (also known as dynamic fixed-point), 
% where a value $x$ is quantized as $x_{\text{quant}} = \left\lfloor x \cdot s_l - z_l \right\rceil$ 
using a \textit{uniform} wordlength\footnote{We use the terms \textit{wordlength} and \textit{bitwidth} interchangeably.} %$b$
across layers.
% and different scale factors and zero points across layers.
% across all layers and with different scale factors 
% $s_l$ and zero points 
% $z_l$ for each layer.
% $l$.
The majority of existing works apply either \textit{1)}~quantization to already trained full-precision models, followed by a retraining step to fine-tune the weights~\cite{angeleye2018tcad,adaquant2021icml}, or \textit{2)}~quantization-aware training to directly obtain low-precision models~\cite{int_only2018cvpr,relaxed_quant2019iclr}. As such, a commonality of both approaches is that they require an expensive training step. 

A third approach that allows for \textit{nonuniform} per-layer wordlength are mixed-precision schemes, such as HAQ~\cite{wang2019haq} and HAWQ~\cite{hawqv22020neurips}. However, both HAQ and HAWQ impose an excessive computational overhead by relying on reinforcement learning and a multi-stage retraining process, respectively. More importantly, both are tailored for classification DNNs.

% However, HAQ imposes an excessive computational overhead by relying on reinforcement learning and the resulting quantized model requires highly customized accelerators to achieve performance gains. More importantly, both HAQ and HAWQ are tailored for classification DNNs.

% \stelios{Discuss HAQ and HAWQ as 3) in the list. This family of techniques is an exception that also adapts the bitwidth of each layer. However, it is tailored for classification DNNs (this is the most important issue), it requires training and it requires exotic specialised accelerators (e.g. with bit-serial processing units) in order to achieve performance results.}

% At the same time, 
Although the aforementioned quantization approaches have been successfully applied on \textit{classification} DNNs
with minimal accuracy loss,
% without incurring significant accuracy loss~\cite{int_only2018cvpr,outlier_quant2018isca}, 
they do not generalize to SR models, as they often lead to a catastrophic drop in visual quality~\cite{mobisr2019mobicom,nemo2020mobicom,PAMSQS}, as shown in Fig.~\ref{fig:qualitative}.
This is primarily due to the removal of Batch Normalization (BN) layers from recent SR models~\cite{RCAN,tpsr2020eccv,IMDN} as they were shown to severely restrict their representational power~\cite{EDSR}. 
In turn, the absence of BN leads to significant variability in the dynamic range of activations, making the direct utilization of existing
quantization methods futile~\cite{int_only2018cvpr} or requiring expensive architectural modifications and retraining~\cite{efficientbnnsrMa2019,bnnsrXin2020,PAMSQS}.

With the integration of low-precision NPUs in smartphones, there is an emerging need for novel quantization methods that are particularly crafted for on-device SR in order to combine high quality with efficiency. 
In this context, 
our \tool framework introduces novel \textit{post-training} techniques that closely approach the quality of full-precision models, leaving \textit{little room for improvement} through expensive retraining. 
In addition, \tool
can be applied complementarily on models trained in a quantization-aware manner.

% \textbf{Challenges and Opportunities of NPUs.}
\subsection{Challenges and Opportunities of NPUs}

Designed explicitly for DNN workloads, mobile NPUs typically rely on low-precision processing units, employing 16- or 8-bit fixed-point arithmetic~\cite{embench2019emdl,ai_benchmark2019iccvw}. 
Despite the potential processing benefits and although such narrow precision has been used effectively for classification DNNs~\cite{int_only2018cvpr}, quantized SR models suffer excessive quality drops compared to their full-precision versions (Fig.~\ref{fig:qualitative}), making NPU execution prohibitive.

Nonetheless, recent hardware advances have led to NPUs that support \textit{multiple arithmetic precisions}.
% ~\cite{snpe,arm_ethos_npu,huawei_npu2019hotchips}. 
Such examples are Hexagon 698 on Qualcomm Snapdragon 865 (SDM865)~\cite{snpe}, Arm Ethos~\cite{arm_ethos_npu} and MediaTek AI processing unit (APU)~\cite{mediatek_apu2020isscc}, all supporting two precision modes: 8-bit activations and weights (INT8) or 16-bit activations and 8-bit weights (A16W8). 
In spite of the new opportunities of these hardware architectures, existing deployment methods fail to exploit them, leading to \textit{1)}~fast but low-quality execution in INT8 - due to the quantization-induced error, \textit{2)}~higher quality but slow execution in A16W8 - close to 2$\times$ slower than INT8, as shown in \S\ref{sec:eval_baselines}), or \textit{3)}~slow \textit{and} low-quality execution in A16W8 for models where even 16 bits do not suffice - which is often the case for SR models. 
Our work pushes the boundaries of what is possible in terms of mapping SR models to NPUs, yielding fast and high-quality designs that fully utilize their multi-precision capabilities.

\vspace{-0.25cm}
\section{\tool Overview}
\label{sec:nawq_overview}
% \vspace{-0.1cm}

Towards addressing the shortcomings of existing mobile SR systems, we propose \tool, an NPU-centric framework that maximizes the efficiency of on-device SR.
\tool leverages the fact that different parts of SR neural architectures have nonuniform precision needs, in order to partition the execution 
across the
NPU's heterogeneous units.  
With SR models deployed across a broad range of use-cases, \tool is in a unique position to enhance the performance of a wide range of visual-content mobile
applications.

% \tool re-examines several concepts related to precision quantization and scheduling, and introduces novel offline and run-time techniques to boost the performance of SR on mobile SoCs.
% Through \tool, we show that we do not need to compromise the visual quality of visual-content applications or penalize their responsiveness in order to provide fast and high-quality on-device super-resolution.
% Instead, \textit{performance can be maximized by means of a smarter utilization of the heterogeneous processing units within mobile NPUs and a quantization-scheduling co-design at the layer level of SR DNNs.} To this end, \tool: 
% \blue{\textit{1)}~introduces a multi-wordlength quantization paradigm that allows the usage of different bitwidth per SR DNN layer; \textit{2)}~co-optimizes the hybrid-precision quantization and the scheduling of layers on the NPU's heterogeneous units; and \textit{3)}~utilizes adaptive arithmetic precision to selectively equip the layers with wider representational power, leading to improved visual quality, despite the fixed-point processing. In this manner, \tool improves the visual quality beyond what was previously possible on low-precision NPUs.}

\begin{figure}[t]
    \centering
    % \vspace{-0.1cm}
    {
    \includegraphics[trim={0cm 6cm 0cm 1cm},clip,width=0.5\textwidth]{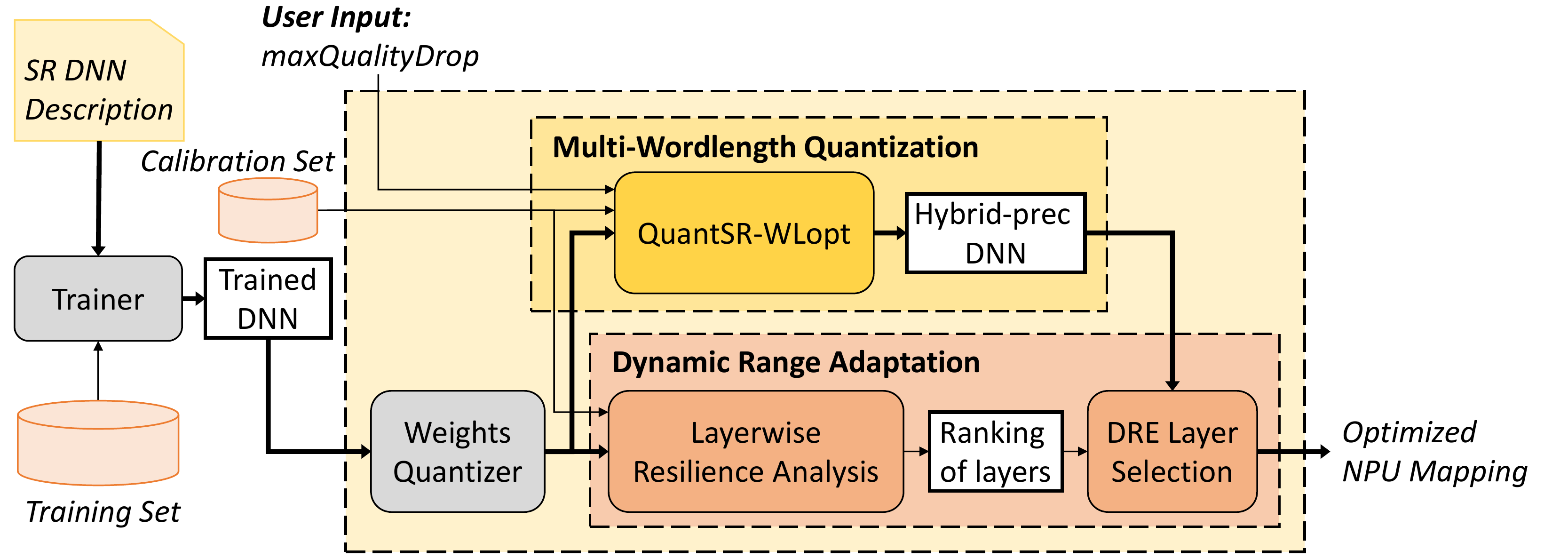}
    \vspace{0.2cm}
    }
    \caption{Overview of \method's offline flow.}
    \label{fig:proc_flow}
    \vspace{-0.45cm}
\end{figure}

\textbf{Offline Flow.}
Fig.~\ref{fig:proc_flow} shows \tool's offline flow. 
The framework is supplied with a trained SR DNN and a quality drop tolerance using an image distortion metric.
As a first step, the \textit{Weights Quantizer} analyses the dynamic ranges of the model's weights in each layer and accordingly reduces their precision to 8 bits, using suitable scale factors. 
Next, the \textit{Multi-Wordlength Quantizer} (\S \ref{sec:multiwl_sr_dnns}) considers the NPU-supported bitwidths and determines the wordlength for the activations of each layer, allowing for different bitwidths across layers. 
The output of this stage is a quantized \textit{hybrid-precision} DNN. 
At this stage, the user-supplied calibration set is used to find the least computationally costly hybrid-precision DNN that meets the user's quality constraint.

As a next step, the weights-quantized DNN is passed to the \textit{Dynamic Range Adaptation} module (\S \ref{sec:dre}). This module is responsible for deciding which layers will \textit{not} use the quantization \textit{scale factors} that the \textit{Multi-Wordlength Quantizer} selected. Instead, these layers derive their scale factors \textit{at run time} by examining the dynamic range of the input activations tensor and quantizing them on-the-fly. We refer to this technique as run-time \textit{dynamic range estimation} (DRE) and determine the DRE layers using
the \textit{DRE Layer Selection} module 
% (\S \ref{sec:dre_layer_sel}) 
based on a \textit{Layerwise Resilience Analysis},
% (\S \ref{sec:lra}), 
which assesses the resilience of each layer to low precision. 
Overall, given the user-defined quality drop tolerance, \tool generates a \textit{DRE-augmented hybrid-precision} model together with an \textit{execution schedule}, tailored for the NPU of the target mobile device and content.

\begin{figure}[t]
    % \vspace{-0.1cm}
    \centering
    % \fbox
    {
    \includegraphics[trim={4cm 7.5cm 6.25cm 0.5cm},clip,width=0.45\textwidth]{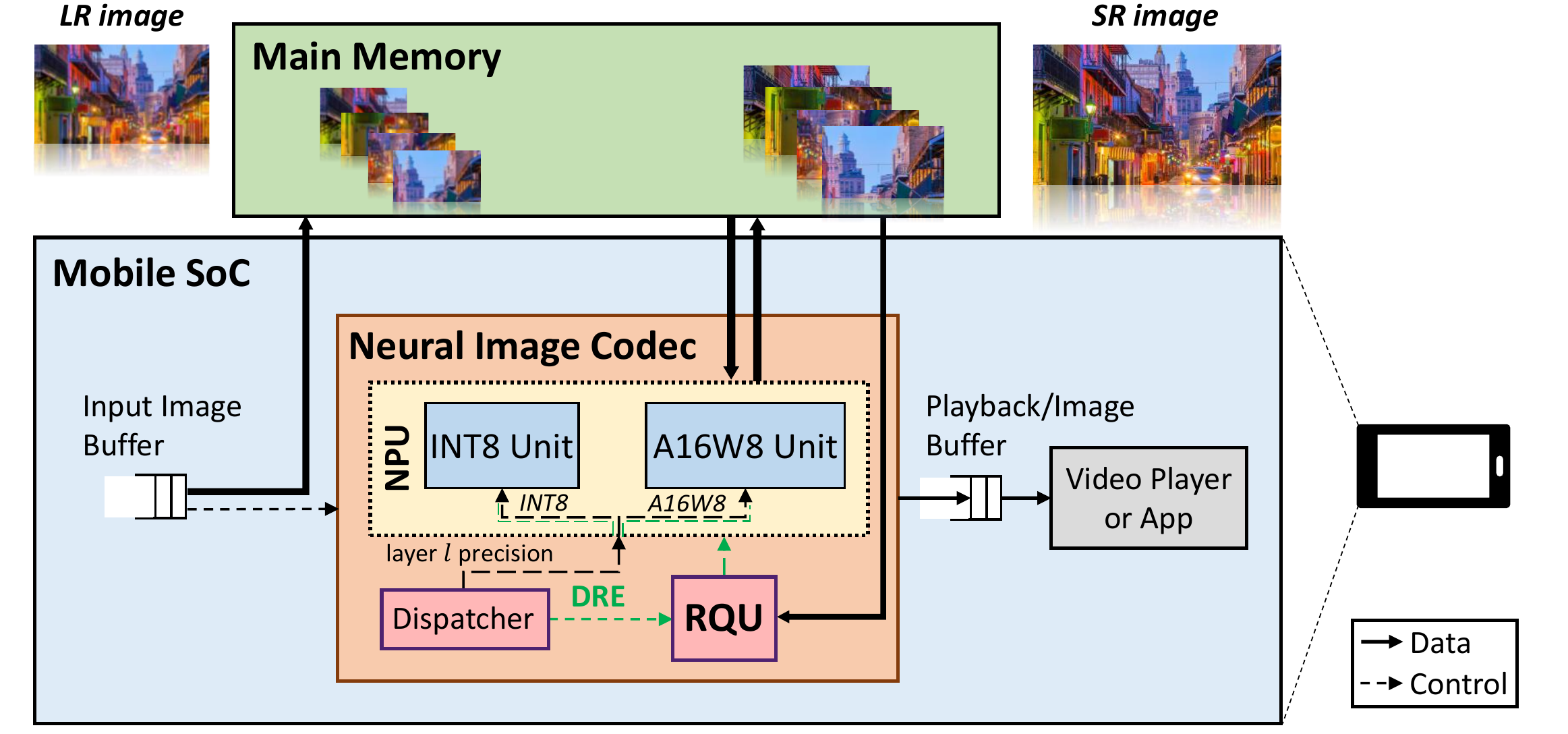}
    \vspace{-0.1cm}
    }
    \caption{\method's runtime architecture.}
    \label{fig:sys_arch}
    \vspace{-0.45cm}
\end{figure}

\textbf{Runtime Architecture.}
\label{sec:runtime_arch}
Fig.~\ref{fig:sys_arch} depicts the architecture of \tool upon deployment. The process is triggered when LR images arrive at the \textit{Input Image Buffer}. 
These are passed in a per-image manner to the \textit{Neural Image Codec} (\S \ref{sec:neural_codec}), which is responsible for their upscaling.
The \textit{Dispatcher}, already hosting the \tool's hybrid-precision model and its associated execution schedule, 
schedules the processing of the input images on the NPU. As such, each layer is executed either on the INT8 or the A16W8 unit. %as dictated by \tool. 
If DRE is selected, the layer's input activations tensor is redirected to the \textit{Runtime Quantization Unit (RQU)}, which in turn quantizes it based on its actual dynamic range and then feeds it to the appropriate unit.
Finally, the processed images are passed to the \textit{Playback/Image Buffer}.

% Fig.~\ref{fig:sys_arch} depicts the architecture of \tool upon deployment. The process is triggered when LR images arrive at the \textit{Input Image Buffer}. 
% These are passed in a per-image manner to the \textit{Neural Image Codec} (\S \ref{sec:neural_codec}), which is responsible for their upscaling.
% The \textit{Dispatcher} (\S \ref{sec:dispatcher}), already hosting the \tool's hybrid-precision model and its associated execution schedule, 
% schedules the processing of the input images on the NPU. As such, each layer is executed either on the INT8 or the A16W8 unit. %as dictated by \tool. 
% If DRE is selected, the layer's input activations tensor is redirected to the \textit{Runtime Quantization Unit (RQU)} (\S \ref{sec:rqu}), which in turn quantizes it based on its actual dynamic range and then feeds it to the appropriate unit.
% Finally, the processed images are passed to the \textit{Playback/Image Buffer}.

\vspace{-0.25cm}
\section{Design of \tool}
\label{sec:nawq_design}
% \vspace{-0.1cm}

% \tool is designed to deliver fast and efficient on-device super-resolution under visual quality requirements.
In this section, we detail how \tool leverages the heterogeneous processing units of mobile NPUs through hybrid-precision execution and formally define the optimization problem that jointly decides the quantization and mapping of DNN layers to the NPU resources. Moreover, we describe the runtime components of \tool and the associated optimizations that ensure efficient and high-performance integration into commodity mobile devices.

% \vspace{-0.3cm}
\vspace{-0.5em}
\subsection{Multiple Wordlengths for Mobile SR}
\label{sec:multiwl_sr_dnns}
% \vspace{-0.1cm}

% Traditional mobile implementations of DNNs employ either floating- or fixed-point representations. While CPUs and GPUs commonly use floating-point arithmetic, DSPs and NPUs typically adopt fixed-point data types. This allows DSPs and NPUs to consume less area on a mobile SoC, leading to more energy-efficient execution and occasionally to higher performance than the use of floating-point. In modern mobile platforms, fixed-point DSPs and NPUs are generally well-known to be more efficient than their floating-point counterparts for most DNN algorithms~\cite{ai_benchmark2019iccvw,embench2019emdl,almeida2021smart,facebook2019hpca}. %fastva2020infocom}.

Traditional mobile implementations of DNNs commonly employ a single uniform wordlength across all computations, with either floating-point arithmetic on CPUs and GPUs or fixed-point on DSPs and NPUs. 
This is a result of targeting pre-designed processing units, such as a CPU's FP32 or a DSP's INT8 units. Nevertheless, the latest NPUs can help us overcome this restriction for two reasons.
At the hardware level, modern NPUs either host heterogeneous processing units that support different arithmetic precision, \textit{e.g.}~the 8-bit HVX and A16W8 HTA units on the Hexagon 698 NPU, or provide precision-configurable units, \textit{e.g.}~Samsung S21's NPU~\cite{samsung_npu2021isca}.
This property allows the optimization of the DNN execution so that different operations are performed using different precision. At the algorithmic level, we can design methodologies that allow the customization 
of each operation's precision, shaping the per-operation wordlength to the requirements of the DNN algorithm.
% (\S\ref{sec:wlopt} and \ref{sec:dre}). 

Together, these optimization opportunities point to an alternative design paradigm, which we name \textit{hybrid-precision}.
This implementation style introduces a multiple-wordlength approach and inherits the speed and energy advantages of fixed-point arithmetic. However, by allowing each operation in the DNN to be encoded with a different wordlength, the design degrees of freedom are significantly increased.

To comply with the widely adopted practice of applying 8-bit quantization on the weights of a model 
and with the NPU trend of supporting only 8-bit weights~\cite{snpe,arm_ethos_npu}, we quantize the weights using 8 bits across all layers \red{(line 1 in Alg.~\ref{alg:static_wl_opt_refined} and Weights Quantizer in Fig.~\ref{fig:proc_flow}}), and tailor our hybrid-precision method to the activations. We first define the \textit{granularity} at which different wordlengths can be applied. In \tool, we opt for a \textit{layerwise} parametrization. This approach ensures the efficient utilization of the underlying hardware: the quantization step prior to execution has to be amortized across several computations, which is achieved by the compute-intensive convolution or matrix operations of a DNN layer. Finer granularity, such as allowing for different wordlength per channel, would incur significant overhead due to the low computation-to-quantization ratio. 
% , counteracting the benefits of our approach.

% Next, we discuss our hybrid-precision method, focusing on: \textit{i)}~the quantization strategy that specifies how a given tensor is quantized %(\S\ref{sec:quant_scheme_1}) 
% and \textit{ii)}~the wordlength optimization algorithm that decides the bitwidth of each layer. 
% (\S\ref{sec:wlopt}). 

% \vspace{-0.2cm}
% \subsubsection{Hybrid-Precision Quantization Strategy}
% \label{sec:quant_scheme_1}
% \vspace{-0.1cm}

\textbf{Hybrid-Precision Quantization Strategy.}
To implement multi-wordlength DNNs, a hybrid-precision quantization strategy needs to be defined. The proposed strategy utilizes different wordlength $b_l$, scale factor $s_l$ and zero point $z_l$ for each layer $l$, such that a value $x$ is quantized to a $b$-bit integer $x_{\text{quant}}$ as $x_{\text{quant}} = \left\lfloor x \cdot s_l - z_l \right\rceil$. 
To introduce different wordlengths among layers, quantization is performed such that all values within each activations tensor at the input of each layer have a single wordlength, scale factor and zero point. As such, the quantization configuration, $q_l$, for the $l$-th layer is given by $q_l = \left< b_l, s_l, z_l \right> ~ \forall l \in \mathcal{L}$,
where $\mathcal{L}$ is the set of layers in the given DNN.
Furthermore, the scale factor $s_l$ and zero point $z_l$ are derived based on the estimated dynamic range of the activations tensor $\mathbf{x}$ as
% \vspace{-0.2cm}
\begin{small}
\begin{equation}
    s_l = \frac{(2^{b_l} - 1)}{\widehat{\mathbf{x}}_{\text{max}} - \widehat{\mathbf{x}}_{\text{min}}} \quad,  \qquad  z_l = \left\lfloor s_l \cdot \widehat{\mathbf{x}}_{\text{min}} \right\rceil \label{eq:scale_factor}
\end{equation}
\end{small}
% \vspace{-0.5cm}

\noindent
where $\widehat{\mathbf{x}}_{\{\text{max},\text{min}\}}$ are estimates of the max/min values in $\mathbf{x}$, derived by processing a dataset that is representative of the target task. We refer to this set as the calibration set.

% \vspace{-0.2cm}
% \subsubsection{Hybrid-Precision Wordlength Optimization}
% \label{sec:wlopt}
% \vspace{-0.1cm}

\textbf{Hybrid-Precision Wordlength Optimization.}
Given a DNN $m$ with $|\mathcal{L}|$ layers, we define a wordlength $b_l$ for each layer $l$, referred to collectively as the vector $\mathbf{b}$.
We further denote by $m(\mathbf{b})$ a model quantized with hybrid precision across its layers as dictated by $\mathbf{b}$.  
Let $\epsilon$ be the user-specified maximum allowable drop on average quality, which can be quantified using the peak signal-to-noise ratio (PSNR) image reconstruction metric, denoted by $\mathbb{E}(Q(m(\mathbf{b})))$. Given a cost estimator $T(m(\mathbf{b}))$ (\textit{e.g.}~latency estimate or FLOPs), we pose the following 
constrained 
optimization problem
\vspace{-0.15cm}
\begin{small}
\begin{flalign}
    &\min\limits_{\mathbf{b}} T(m(\mathbf{b})) \quad \text{subject to} \label{eq:opt_problem} \\
    &\forall l \in \mathcal{L}: b_l \in \mathcal{W} \quad \text{and} \quad \mathbb{E}\Big(Q\big(m(\mathbf{b})\big)\Big) - \mathbb{E}\Big(Q\big(m(\mathbf{u})\big)\Big) \le \epsilon \nonumber
\end{flalign}
\end{small}
\vspace{-0.6cm}

\noindent
where $\mathcal{W}$ is the candidate wordlength set and $\mathbf{u}$ is the uniform wordlength vector that assigns 32 bits to all layers. The scale factor $s_l$ and zero point $z_l$ are analytically derived as per Eq.~(\ref{eq:scale_factor}) and hence are implicitly co-optimized with the selection of $b_l$. Thus, we omit them from Eq.~(\ref{eq:opt_problem}).

The optimization considers the supported bitwidths of the underlying NPU (\textit{e.g.}~$\mathcal{W}=\{ 8, 16\}$ for SDM865) and aims to find the wordlengths and scale factors of all layers that minimize the execution cost of an SR DNN on the NPU, subject to the given quality constraints.
To capture the execution cost on the specialized hardware of NPUs, we adopt a variation of the number of bit operations (BOPs) metric as our cost estimator $T$~\cite{uniq2021tocs,relaxed_quant2019iclr}. Our metric weights each operation with a cost based on the number of bytes used. Specifically, operations performed in 32, 16, and 8 bits are assigned a cost of 4, 2 and 1, respectively, reflecting the runtime and memory differences among the different bitwidths. Hence, given a model $m$ and a wordlength vector $\mathbf{b}$, GetBOPs($m(\mathbf{b}$)) returns the total cost of executing $m$ by considering each layer's number of operations and assigned wordlength ($b_l$).

The per-layer wordlength selection can be cast as a search problem aiming to achieve peak processing speed by selecting suitable bitwidths. For an SR DNN with $|\mathcal{L}|$ layers and $|\mathcal{W}|$ candidate bitwidths, the total number of candidate hybrid-precision configurations is $|\mathcal{W}|^{|\mathcal{L}|}$. With an increase in either the depth of a DNN or the number of available bitwidths, an exhaustive enumeration rapidly becomes intractable. In real-world deployments, although NPUs currently support up to two bitwidths, \textit{e.g.}~8 or 16 bits, state-of-the-art SR DNNs reach significant depths, ranging from 33 layers for the lightweight TPSR model~\cite{tpsr2020eccv} and hence 8 billion design points, up to more than 1500 layers for RCAN~\cite{rcan2018eccv} and $2^{1500}$ design points. As a result, the combinatorial scaling of the design space size and the large depth of SR DNNs prohibit optimization by means of enumeration.

\textbf{QuantSR-WLopt.}
In this context, we propose QuantSR-WLopt, a heuristic method to obtain a solution in the non-convex design space. 
The key principle behind QuantSR-WLopt is a cost-prioritizing strategy that applies more aggressive quantization to the most FLOPs-heavy layers first, through an efficient single-shot wordlength adaptation, \textit{i.e.}~by modifying the wordlength of each layer only once.

% NPU-centric Hybrid-Precision Wordlength Optimization (QuantSR-WLopt)
\setlength{\textfloatsep}{0pt}
\SetArgSty{textnormal}
\begin{algorithm}[!t]
    \setcounter{AlgoLine}{0}
    \scriptsize
    \SetAlgoLined
    \LinesNumbered
    \DontPrintSemicolon
    \KwIn{DNN $m$ with layers $\mathcal{L}$, Wordlengths set $\mathcal{W}=\{8,16\}$}
    \nonl
    \myinput{Calibration set $\mathcal{D}_{\text{calib}}$}
    
    \nonl \myinput{Reference quality $q_{\text{ref}}$ (PSNR in dB or SSIM in $[-1,1]$)}
    
    \nonl
    \myinput{Quality drop tolerance $\epsilon$}
    
    \KwOut{Optimized wordlength vector $\mathbf{b}^{\text{sel}} \in \mathcal{W}^{|\mathcal{L}|}$}

    \red{$m \leftarrow$ WeightsQuantizer($m$, 8)} \Comment*[f]{\scriptsize \textrm{\red{Quantize weights to 8 bits}}}
    
    $\mathbf{u} \leftarrow$ uniform wordlength (in our case 16 bits)
    
    $\mathbf{b}^{\text{sel}} \leftarrow \mathbf{u}$
    
    InitScales\&ZeroPoints($m(\mathbf{b})$, $\mathcal{D}_{\text{calib}}$)
    
    $c^{\text{bops}}_{\text{total}}, \mathbf{c}^{\text{bops}}_{\text{layers}} \leftarrow \text{GetBOPs}(m(\mathbf{b}))$ \Comment*[f]{\scriptsize \textrm{Initial cost}}
    
    $\mathbf{c}^{\text{sorted}}_{\text{layers}}, \mathcal{L}^{\text{sorted}}_{\text{bops}} \leftarrow \text{SortDescending}(\mathbf{c}^{\text{bops}}_{\text{layers}})$ 
    
    \ForEach(\Comment*[f]{\scriptsize \textrm{Single-shot pass through the layers}}){$l$ \textbf{in} $\mathcal{L}^{\text{sorted}}_{\text{bops}}$}{
        $\mathbf{b} \leftarrow \mathbf{b}^{\text{sel}}$ 
        
        $b_l \leftarrow 8$ 
        
        UpdateScale\&ZeroPoint($b_l$) \Comment*[f]{\scriptsize \textrm{Using Eq.~(\ref{eq:scale_factor})}}
        
        $q \leftarrow \text{GetQuality}(m(\mathbf{b}), \mathcal{D}_{\text{calib}})$ 
        
        $c^{\text{bops}} \leftarrow \text{GetBOPs}(m(\mathbf{b}))$
        
        \If(\Comment*[f]{\scriptsize \textrm{Quality constraint}}){$q_{\text{ref}} - q \le \epsilon$}{
            $\mathbf{b}^{\text{sel}}_{l} \leftarrow 8$, \quad
            $q_{\text{best}} \leftarrow q$, \quad $c^{\text{bops}}_{\text{min}} \leftarrow c^{\text{bops}}$ 
        }
    }
    \caption{\footnotesize Wordlength Optimization (\textbf{QuantSR-WLopt})}
    \label{alg:static_wl_opt_refined}
\end{algorithm}

With reference to Algorithm~\ref{alg:static_wl_opt_refined} and with a running example of $\mathcal{W}=\{8, 16\}$, QuantSR-WLopt first quantizes all layers with the same uniform high precision (\textit{e.g.}~16 bits) (lines~1-3) and sorts them with respect to the amount of BOPs (lines~4-5).
Next, the algorithm iterates \textit{once} along the depth of the DNN and sets the wordlength of the $l$-th layer to 8 bits (line 8). By passing through the calibration set, the achieved quality $q$ is calculated (line~10), together with the new cost (line~11). If the current quality satisfies the constraint, layer $l$ is kept to 8 bits; else it is reverted back to 16 bits to recover the lost quality (lines~12-14).

QuantSR-WLopt exhibits a number of crucial properties. With respect to complexity, it scales linearly with the number of layers $|\mathcal{L}|$ as each layer is examined only once. With respect to execution cost, 
% by prioritizing the higher-cost layers, 
QuantSR-WLopt's cost-aware criterion ensures that a less costly layer is never quantized to lower precision at the expense of a heavier layer. Hence, it prioritizes the quantization of layers that will have a larger impact on minimizing the runtime. With respect to quality, the algorithm guarantees by design the return of a configuration that meets the quality constraint, if and only if such a design exists in the design space. As such, the upper bound in quality is given by $m(\mathbf{b}^{\text{max}})$ where $b_l^{\text{max}} = \max(\mathcal{W})$ for all $l \in \mathcal{L}$. 
Thus, to address cases where the upper bound in quality is not satisfactory, we introduce a new design dimension in the quantization scheme by deciding whether to fix or dynamically determine the scale factor and zero point of each layer. We discuss this in the following section.

% \vspace{-0.3cm}
\vspace{-0.5em}
\subsection{Dynamic Range Adaptation}
\label{sec:dre}
% \vspace{-0.1cm}

As described in Section~\ref{sec:multiwl_sr_dnns}, the A16W8 mode constitutes our scheme's upper bound in attainable visual quality. However, there are cases where A16W8 fails to satisfy the constraint of Eq.~(\ref{eq:opt_problem}). As such, current NPU mappings often fail to reach acceptable quality, especially when targeting efficient SR models. This has led to existing works either partially using the NPU~\cite{mobisr2019mobicom} or avoiding it altogether~\cite{nemo2020mobicom,splitsr2021imwut}.

To push the quality of NPU-based SR beyond what was previously attainable,
while sustaining the processing benefits of hybrid-precision execution, \tool introduces a new design dimension to the quantization strategy, which we name \textit{dynamic range estimation} (DRE). DRE adapts the scale factor and zero point of an activations tensor \textit{at run time}, based on the \textit{actual} range of its values for the \textit{particular} input sample. 
This technique overcomes the limitations of existing works, where the values of $s_l$ and $z_l$ are statically derived prior to deployment and remain fixed at run time.  
The primary limitation that leads to degraded output quality is manifested in cases where the estimated dynamic range does not capture the actual encountered range of an input. In these cases, the statically determined precision underutilizes the representation range of the selected wordlength, leading to excessive numerical error and, in turn, quality drop.
Instead, DRE adapts the scale factor and zero point in an input-dependent manner, occupying the full range of values for the activations of the current input.

With this scheme, we formulate the new quantization method for each layer as $q_l = \left< b_l, s_l, z_l, d_l \right> ~ \forall l \in \mathcal{L}$,
where $d_l \in \{0,1\}$ indicates whether DRE is applied on layer $l$. When $d_l$ is $1$ and DRE is enabled, the actual dynamic range of the input activations tensor $\mathbf{x}$ is first calculated and the scale factor $s_l$ and zero point $z_l$ are derived on-the-fly as per Eq.~(\ref{eq:scale_factor}), by substituting the statically determined estimates at the denominator with the actual values, ${\mathbf{x}}_{\text{max}}$ and ${\mathbf{x}}_{\text{min}}$.

% Layerwise Resilience Analysis (LRA) algorithm
\setlength{\textfloatsep}{0pt}
\SetArgSty{textnormal}
\begin{algorithm}[!t]
    \setcounter{AlgoLine}{0}
    \scriptsize
    \SetAlgoLined
    \LinesNumbered
    \DontPrintSemicolon
    \KwIn{DNN $m$ with layers $\mathcal{L}$, Wordlengths set $\mathcal{W}=\{8,16\}$}
    % \nonl
    % \myinput{Wordlengths set $\mathcal{W}=\{8,16\}$}
    \nonl
    \myinput{Quality drop $q_{\text{w-quant}}^{\text{drop}}$  of 8-bit weights-quantized DNN}
    
    \nonl
    \myinput{Reference quality $q_{\text{ref}}$ (PSNR in dB or SSIM in $[-1,1]$)}
    
    \KwOut{Sorted layers with respect to quality drop $\mathcal{L}^{\text{sorted}}_{\text{drop}}$}
    \nonl
    
    \myoutput{Sorted layerwise quality drops $\mathbf{q}^{\text{drop}}_{\text{sorted}}$}
    % \nonl\;
    
    $q_{\text{ref}} \leftarrow q_{\text{ref}} -     q_{\text{w-quant}}^{\text{drop}}$ \Comment*[f]{\scriptsize \textrm{Remove quality drop due to 8-bit weights}}
    
    $u \leftarrow$ uniform wordlength (in our case 16 bits) 
    
    \ForEach( \Comment*[f]{\scriptsize \textrm{for each layer}}){$l$ \textbf{in} $\mathcal{L}$} {
        $\mathbf{b} \leftarrow \mathbf{u}$
        
        $b_l \leftarrow 8$ \Comment*[f]{\scriptsize \textrm{\red{Set bitwidth for the $l$-th layer's activations to 8 bits}}}

        $q \leftarrow \text{GetQuality}(m(\mathbf{b}))$
        
        $\mathbf{q}^{\text{drop}}_l \leftarrow q_{\text{ref}} - q$
    }
    $\mathbf{q}^{\text{drop}}_\text{sorted}, \mathcal{L}^{\text{sorted}}_{\text{drop}} \leftarrow \text{SortDescending}(\mathbf{q}^{\text{drop}})$ 

    \caption{\footnotesize Layerwise Resilience Analysis (\textbf{LRA})}
    \label{alg:lra}
\end{algorithm}

The advantages of DRE come at a cost:
% When DRE is applied on a given layer, the additional
the computational overhead of finding the actual range (\textit{i.e.}~min/max values) of the activations tensor and computing the new scale factor and zero point has to be taken into account. 
In other words, applying DRE across all layers in a brute-force manner can lead to excessive latency and thus negate its benefits. 
Hence, to effectively utilize DRE, we have to devise a method of: \textit{i)}~quantifying the resilience of each layer to low precision, %(\S\ref{sec:lra}), 
and \textit{ii)}~an algorithm that leverages this information to \textit{selectively} apply DRE to a subset of the DNN's layers. % (\S\ref{sec:dre_layer_sel}). 
% To this end, we next present the \textit{Layerwise Resilience Analysis} (LRA) and \textit{DRE Layer Selection} methods that address each respective problem in \tool.

% \vspace{-0.2cm}
% \subsubsection{Layerwise Resilience Analysis}
% \label{sec:lra}
% \vspace{-0.1cm}

\textbf{Layerwise Resilience Analysis.}
Algorithm~\ref{alg:lra} presents our technique for estimating each layer's resilience to reduced precision.
The core idea behind LRA is to isolate each layer's contribution to the quality drop of a quantized model. 
As the weights are already 8 bits (\S\ref{sec:multiwl_sr_dnns}), we first subtract the PSNR drop caused solely by the weights quantization (line~1). In this manner, any subsequently observed PSNR degradation is due to the activations quantization. 
The algorithm starts by using a uniform higher-precision representation for the activations of all layers (line~2). Next, we iterate through the layers, quantizing each one \textit{individually} to 8 bits and obtaining the associated drop with respect to that of the weight-quantized model (line 7).
Finally, the layers are sorted in a decreasing order of quality drop (line~8).

% DRE Layer Selection algorithm
\setlength{\textfloatsep}{0pt}
\SetArgSty{textnormal}
\begin{algorithm}[!t]
    \setcounter{AlgoLine}{0}
    \scriptsize
    \SetAlgoLined
    \LinesNumbered
    \DontPrintSemicolon
    \KwIn{Hybrid-precision DNN $m_q$ with layers $\mathcal{L}$}
    \nonl
    \myinput{Sorted layers with respect to quality drop $\mathcal{L}^{\text{sorted}}_{\text{drop}}$}
    
    \nonl
    \myinput{Sorted layerwise quality drops $\mathbf{q}^{\text{drop}}_{\text{sorted}}$}
    
    \nonl 
    \myinput{Energy concentration threshold $K \in [0,1]$}
    
    \KwOut{DRE-augmented quantized model $m_q^{\text{DRE}}$}
    % \nonl\;
    
    \For(\Comment*[f]{\scriptsize \textrm{loop through sorted layers}}){$l \gets 0$ \textbf{to} $|\mathcal{L}|-1$} {
        $\mathbf{E}_l \leftarrow \sum\limits_{i=0}^{l}|\mathbf{q}_{\text{sorted},l}^{\text{drop}}|^2$ \Comment*[f]{\scriptsize \textrm{Energy concentration up to layer $l$}}
    }
    
    \For(\Comment*[f]{\scriptsize \textrm{loop through sorted layers}}){$l \gets 0$ \textbf{to} $|\mathcal{L}|-1$} {
       \If(\Comment*[f]{\scriptsize \textrm{Energy constraint}}){$ \mathbf{E}_l ~/~ \mathbf{E}_{|\mathcal{L}|-1} \leq K $}{
            $\mathcal{L}_{\text{DRE}} \leftarrow \text{Append}\big(\mathcal{L}^{\text{sorted}}_{\text{drop}}(l)\big)$
        }  
    }
    $m_q^{\text{DRE}} \leftarrow \text{AddDRE}(m, \mathcal{L}_{\text{DRE}})$ \Comment*[f]{\scriptsize \textrm{Use DRE on the selected layers}} \\ 
    \caption{\footnotesize DRE Layer Selection}
    \label{alg:nawq-sr}
\end{algorithm}

% \vspace{-0.2cm}
% \subsubsection{DRE Layer Selection}
% \label{sec:dre_layer_sel}
% \vspace{-0.1cm}

\textbf{DRE Layer Selection.}
After selecting the highest performing bitwidths via QuantSR-WLopt and estimating the layerwise resilience to quantization through LRA, \tool picks a subset of layers, to have their scale factors and zero points computed at run time based on their actual dynamic range. 
Algorithm~\ref{alg:nawq-sr} describes this layer selection process. The objective of the algorithm is to recover the visual quality for the layers which exhibit large quality degradation when quantized.
Our key insight is to interpret the layerwise PSNR drop as a discrete signal and adopt the respective signal energy~\cite{signalproc2015book} (line~2) as a criterion to tune the amount of layers that will utilize DRE.
Given the DNN layers ordered by quality drop, the DRE layer selection algorithm first calculates the energy concentration up to each of these layers (lines~1-3). For instance, the energy concentration of a layer $l$ includes the energy concentration of the previous ordered layers ($0$ to $l$-$1$). 
Next, the algorithm selects for DRE all the layers until the first one that meets the requested energy concentration threshold $K$ (lines~4-7).
Threshold $K$ is represented as a fraction of the total energy concentration ($K \in [0,1]$) and allows for enhancing quality at the expense of the extra DRE-induced latency (quantified in \S\ref{sec:eval_neural_img_codec}).
A key property of our method is that the number and selection of layers that use DRE \textit{do not require tuning}; instead, they are adapted automatically based on $K$ and can be nonuniform across different SR DNNs for the same value of $K$.
% , by adapting which and how many layers use DRE. A key property of our method is that the number of layers that use DRE does not require tuning. Instead, it is selected automatically and is nonuniform across different SR DNNs even for the same value of $K$.

% \vspace{-0.3cm}
\vspace{-0.5em}
\subsection{Neural Image Codec}
\label{sec:neural_codec}
% \vspace{-0.1cm}

The Neural Image Codec is responsible for dividing the LR images into fixed-size patches and upscaling them using the target SR DNN through an optimized NPU mapping.

% \vspace{-0.2cm}
% \subsubsection{Dispatcher}
% \label{sec:dispatcher}
% \vspace{-0.1cm}

\textbf{Dispatcher.}
To guide the on-device execution, the Neural Image Codec introduces a dispatcher that, given the per-layer quantization configuration $q_l$, schedules execution to the appropriate hardware processor of the NPU, using the specified bitwidth, scale factor and zero point. 
To ensure efficient execution, this process is performed in a number of  steps.
First, the dispatcher adopts a partitioning strategy to reduce the communication between the codec components and the target processors. Specifically, the dispatcher partitions the DNN into groups of consecutive layers based on their target bitwidth (\textit{e.g.}~INT8 or A16W8) and range estimation technique ($d_l$), scheduling execution on a per-partition basis. As such, the scheduling of consecutive layers that need to interact with the same components is coalesced, amortizing the cost of communication between components.

Second, the dispatcher considers the requested range estimation technique ($d_l$). Partitions without DRE can be executed without additional supervision using the supplied scale factors and zero points. The remaining partitions are monitored by the RQU to adjust the per-layer scaling factors and zero points at run time, as detailed in the next section. 

Finally, the dispatcher coordinates with the NPU executor to perform inference on a target processor (\textit{e.g.}~either HVX or HTA in SDM865's NPU) that supports the requested partition's bitwidth.
We note that while the DNN partitions are represented with distinct bitwidths, their weights are always in 8 bits and, hence, only activations are quantized on-the-fly.
As such, \tool shares the weights between the activation wordlength representations \red{by storing a single 8-bit copy in memory} and thus incurs no extra memory cost for supporting both INT8 and A16W8.
% As a last step, the resulting upscaled SR patches are passed to the Playback/Image Buffer to be consumed by the application.

Many commercial NPUs already provide either dedicated processors or extra cores for orchestrating execution where \tool's dispatcher can be integrated. Such instances are the Q6 processor in QC's AI processor~\cite{snpe}, or the NPU controller (NPUC) in the latest Samsung Exynos chipsets~\cite{exynos_npu2019isscc,samsung_npu2021isca}.
By executing on a separate processor, \tool's dispatcher and the partitioned inference can be performed in parallel in a pipelined fashion, thus sustaining high utilization of the NPU resources, while requiring no access to the resources of the main CPU and improving the overall efficiency.

% \vspace{-0.2cm}
% \subsubsection{Runtime Quantization Unit}
% \label{sec:rqu}
% \vspace{-0.1cm}

\textbf{Runtime Quantization Unit.}
For the partitions that require DRE, the RQU is responsible for estimating the per-layer dynamic range and adapting the respective scale factors and zero points at run time. 
To derive the new scale and zero point values (as detailed in \S\ref{sec:dre}), the RQU captures each layer's input tensors and extracts their range of values (\textit{i.e.}~${\mathbf{x}}_{\text{min}}$ and  ${\mathbf{x}}_{\text{max}}$). Then, the unit proceeds with the computation of the new scale factor and zero point as dictated by Eq.~(\ref{eq:scale_factor}).
The layer's inputs are then quantized using the new computed parameters and fed to the appropriate processing unit for the actual layer execution.

To be deployable without starving the resources of the target mobile device, the RQU has to exhibit low resource usage when invoked. To this end, we first vectorize the max/min operations by dividing the input activations tensor across parallel max/min search tasks and then apply a parallel-reduce operation to obtain the final range.
Moreover, the RQU execution is placed on the same processing unit as the layers' partition at hand, to avoid unnecessary data transfers.
Overall, the use of DRE results in improved quality with minimal overhead as shown in Section~\ref{sec:eval_neural_img_codec}.

% \vspace{-0.2cm}
% \subsection{Additional Optimizations}
% \label{sec:additional_opt}
% \vspace{-0.2cm}
\textbf{Memory-Aware Mapping of Upsampling.}
Modern state-of-the-art SR DNNs employ pixel-shuffle~\cite{ESPCN} for upsampling %the activations feature maps 
to the desired resolution. 
However, due to the limited cache of NPUs~\cite{npu2020mdat,fastva2020infocom,arm_ethos_npu_perf,snpe_limitations} and pixel-shuffle's excessive memory demands, these layers cannot be directly mapped to NPU, leading to runtime errors~\cite{snpe_limitations}, \blue{substitution with less performant blocks~\cite{nemo2020mobicom} or expensive fallback to CPU-based execution}. 
% The source of inefficiency may be primarily attributed to the 6-dimensional intermediate data of the pixel-shuffle operation.
\blue{This may be primarily attributed to the 6-dimensional intermediate data of the pixel-shuffle operation, which, if not manipulated efficiently, significantly affect the memory footprint.}
It is often the case that the NPU executor attempts to partition the tensor by storing each dimension on a separate memory bank, to provide the processing units with parallel access to all dimensions~\cite{snpe_limitations}. Hence, in cases where the tensor dimensions exceed the number of NPU memory banks or the depth of the banks is severely underutilized, the NPU can run out of memory. 
% \blue{As a result, the schedule and order of significantly affects the memory footprint during execution.}

To address this problem, we introduce a data layout transformation \blue{that caps and minimizes the footprint of pixel-shuffle-based upsampling}. 
Our approach restructures the input and activation tensors so that a maximum of four dimensions are used throughout the pixel-shuffling process.

The original pixel-shuffle operation with an upscale factor of $s$ on a tensor $\mathbf{x}\in \mathbb{R}^{1 \times c_{\text{in}} \times h \times w}$ with $c_{\text{in}}$ channels, height $h$ and width $w$ involves the following steps:
{\footnotesize
\begin{enumerate}[leftmargin=*,noitemsep,topsep=1pt]
    \item \texttt{Reshape} 4D tensor $\mathbf{x}$ into a 6D tensor of shape: {$1 \times c_{\text{out}} \times s \times s \times h \times w$}
    \item \texttt{Permute} dimensions as: {$1 \times c_{\text{out}} \times h \times s \times w \times s$}
    \item \texttt{Reshape} 6D tensor into final 4D tensor of shape: {$1 \times c_{\text{out}} \times s \cdot h \times s \cdot w$}
\end{enumerate}
}
\noindent
This implementation leads to underutilization of the NPU memory.
Instead, we perform the following steps:
{\footnotesize
\begin{enumerate}[leftmargin=*,noitemsep,topsep=1pt]
    \item \texttt{Reshape} 4D tensor into a 2D tensor of shape:
    $c_{\text{out}} \times s \cdot s \cdot h \cdot w$ 
    \item Extract each of the $c_{\text{out}}$ channels in parallel, producing $c_{\text{out}}$ 1D tensors of size: $s \cdot s \cdot h \cdot w$
    \item \texttt{Reshape} each of the $c_{\text{out}}$ 1D tensors to a 4D tensor of shape:
    $s \times s \times h \times w$
    \item \texttt{Permute} each of the $c_{\text{out}}$ 4D tensors as $h \times s \times w \times s$
    \item \texttt{Reshape} each of the $c_{\text{out}}$ 4D tensors to 2D tensor of shape: $s \cdot h \times s \cdot w$
    \item \texttt{Stack} $c_{\text{out}}$ 2D tensors to form a single 3D tensor of shape: $c_{\text{out}} \times s \cdot h \times s \cdot w$
\end{enumerate}
}
\noindent
In this manner, we never exceed 4D tensors and the memory of the NPU is more fully utilized, enabling the mapping of upsampling layers on the NPU. This technique was crucial in order to enable the full NPU-based execution of SR DNNs and avoid the costly CPU fallback of current deployments.

\vspace{-0.25cm}
\section{Evaluation}
\label{sec:eval}
% \vspace{-0.1cm}

% We evaluate the performance of \tool by assessing its components and comparing with highly optimized status-quo designs and state-of-the-art on-device SR systems.

% \vspace{-0.2cm}
% \subsection{Experimental Setup}
% \label{sec:exp_setup}
% \vspace{-0.2cm}

\textbf{Experimental Setup.}~We target the Qualcomm Snapdragon 865 SoC (SDM865) of a Samsung Galaxy S20. SDM865 comprises an octa-core Kryo 585 CPU, an Adreno 650 GPU and the Hexagon 698 NPU. The NPU integrates a vector processor (HVX) supporting INT8 and a tensor accelerator (HTA) supporting both INT8 and A16W8. We consider {$\mathcal{W}=\{8, 16\}$} as our activations wordlengths and map INT8 to HVX and A16W8 to HTA.
We implemented the \tool's offline components using PyTorch (v1.6) and the runtime components by leveraging the Snapdragon Neural Processing Engine (SNPE v1.47) SDK. \red{To showcase the generality of our system, we further target the Snapdragon 888 SoC (SDM888) and present comparisons against highly optimized baselines in terms of processing speed (\S~\ref{sec:eval_baselines}) and energy efficiency (\S~\ref{sec:eval_energy}). SDM888 is hosted on a Snapdragon 888 Mobile Hardware Development Kit (HDK) and consists of an octa-core Kryo 685 CPU, an Adreno 660 GPU and the Hexagon 780 NPU. The NPU comprises scalar, vector and tensor processing units, which are composable and support both INT8 and A16W8. Unless mentioned otherwise, SDM865 is used for measurements.}
% We provide further details of the \tool parameters in Appendix~\ref{app:nawqsr}.
% regarding the \tool's parameter tuning. %selection for each dataset.
%As detailed in Appendix~\ref{app:nawqsr}, we used a quality drop tolerance ($\epsilon$) of $0.1$ and an energy concentration threshold ($K$) of $0.125$, $0.5$ and $1.0$, for IMDN, TPSR and MobiSR-RCAN, respectively. Details in .

\textbf{SR Models.} 
We target three state-of-the-art models of varying depth, architecture and workload: the lightweight TPSR~\cite{tpsr2020eccv}, the mid-range IMDN~\cite{IMDN}, and MobiSR-RCAN~\cite{mobisr2019mobicom}, an efficient RCAN~\cite{RCAN} variant.
% We provide details on these models' pre-training process in Appendix~\ref{app:training}.
% adopted by MobiSR~\cite{mobisr2019mobicom}. 

% \begin{figure}[t]
%     \vspace{-0.2cm}
%     \centering
%     % \fbox
%     {
%     \includegraphics[trim={2.25cm 2cm 2cm 2cm},clip,width=0.34\textwidth]{figures/sota_bicubic_b100_annotated.pdf}
%     }
%     \vspace{-0.5em}
%     \caption{PSNR achieved by the state-of-the-art models 
%     on the B100 dataset with $\times$4 upscaling in recent years.}
%     \label{fig:sota_bicubic}
% \end{figure}

\textbf{Training Details}
% Throughout the evaluation in Section~\ref{sec:eval} we utilize distinct models.
For TPSR and IMDN, we utilize the pre-trained models as provided by the respective authors. 
For MobiSR-RCAN, we follow the training scheme by Lee~\textit{et al.}~\cite{mobisr2019mobicom} and reproduce the reported results.
Following the common practice of both the SR~\cite{EDSR,ESRGAN,RCAN,IDN,IMDN,xlsr2021cvprw} and mobile~\cite{mobisr2019mobicom,supremo2020tmc,live_streaming2020sigcomm,splitsr2021imwut} communities, all models were trained on DIV2K~\cite{DIV2K}, consisting of 800 diverse-content images of 2K resolution. 
Unless otherwise mentioned, we use an upscaling factor of $\times$4 to compare with previous works.

% \textbf{Training Details.}
% We utilize pre-trained models for TPSR and IMDN, provided by the respective authors. For MobiSR-RCAN, we follow the training scheme in ~\cite{mobisr2019mobicom} and reproduce the reported results.
% Following the common practice of both the SR~\cite{EDSR,ESRGAN,RCAN,IDN,IMDN,xlsr2021cvprw} and mobile~\cite{mobisr2019mobicom,supremo2020tmc,live_streaming2020sigcomm,splitsr2021imwut} communities, all models were trained on DIV2K~\cite{DIV2K}, consisting of 800 diverse-content images of 2K resolution. 
% We use an upscaling factor $\times$4 in order to compare with previous works.

\textbf{Performance Metrics.}
We report both visual quality and latency as evaluation metrics.
\blue{In order to compare with other works,}
% For the first, 
we use the standard SR reconstruction quality metrics: PSNR and structural similarity (SSIM)~\cite{SSIM}. 
% \textbf{PSNR is a logarithmic metric.} As such, seemingly minimal gains of 0.1~dB are significant and perceivably important as shown in the improvement of compute-excessive, non-mobile-friendly, state-of-the-art models (Fig.~\ref{fig:sota_bicubic}).
\blue{We also note that a gain in these quality metrics do not necessarily translate to more visually pleasing images~\cite{Blau2018ThePT} and would like to emphasize that seemingly minimal gains of 0.1~dB can help counteract undesirable artifacts which occur due to quantization. As such, we present qualitative results in Fig.~\ref{fig:qualitative}.}
For processing speed, we report the average latency across 50 runs, with the latency measurements obtained through SNPE's timing utilities. Unless mentioned otherwise, we assume a target high-resolution of 720p.

\textbf{Datasets.}
The evaluation was conducted on the standard SR benchmarks used across a large body of recent mobile SR works~\cite{mobisr2019mobicom,supremo2020tmc,splitsr2021imwut,xlsr2021cvprw}, namely Set5~\cite{Set5}, Set14~\cite{Set14}, B100~\cite{B100}, and Urban100~\cite{Urban100}.
Set5 and Set14 are smaller datasets with 5 and 14 images, respectively, each with different SR challenges. B100 and Urban100, with 100 images each, represent a wider range of natural and urban scenes which might be more representative of SR tasks in the wild. For each benchmark dataset, we use 10\% as our calibration set, sampled randomly with uniform probability. Note that while our calibration set selection performs quite well, further exploration of the optimal calibration set size for each model and dataset can be performed~\cite{adaquant2021icml}.

% \textbf{Datasets.} The evaluation was conducted on the standard SR benchmarks used across a large body of recent mobile SR works~\cite{mobisr2019mobicom,supremo2020tmc,splitsr2021imwut,xlsr2021cvprw}, namely Set5~\cite{Set5}, Set14~\cite{Set14}, B100~\cite{B100}, and Urban100~\cite{Urban100}.
% Set5 and Set14 are smaller datasets with 5 and 14 images, respectively, with different SR challenges, while B100 and Urban100, with 100 images each, represent a wider range of natural and urban scenes which might be more representative of SR tasks in the wild.

\textbf{\tool Parameters}
\tool exposes two parameters used for the exploration of the per-layer wordlengths and for the DRE layer selection -- the quality drop tolerance ($\epsilon$) and the energy concentration threshold ($K$), respectively. 
Unless mentioned otherwise, we use a tolerance $\epsilon$ of $0.1$. 
For the model-dataset pairs where weights quantization (FP32W8 in Table~\ref{tab:psnr}) leads to $\geq$ 0.1 dB PSNR drop with respect to the original model (FP32), the tolerance $\epsilon$ is considered with respect to FP32W8 (bold values in the table).\footnote{FP32W8's drop can be reduced further via more sophisticated weight-quantization methods and thus is orthogonal to this work.}
For the energy concentration threshold, we tune the value of $K$ via grid search for each model-dataset pair. 
As such, $K$ was set to $0.125$, $0.5$ and $1.0$, for IMDN, TPSR and MobiSR-RCAN, respectively.

% \textbf{\tool Parameters.}
% \tool exposes two parameters used for the exploration of the per-layer wordlengths and for the DRE layer selection -- the quality drop tolerance ($\epsilon$) and the energy concentration threshold ($K$), respectively. 
% Unless mentioned otherwise, the tolerance $\epsilon$ is set to $0.1$. For the model-dataset pairs where weights quantization (FP32W8 in Table~\ref{tab:psnr}) leads to $\geq$ 0.1 dB PSNR drop with respect to the original model (FP32), the tolerance $\epsilon$ is considered with respect to FP32W8 (bold values in the table).\footnote{FP32W8's drop can be reduced further via more sophisticated weight-quantization methods and thus is orthogonal to this work.}
% For the energy concentration threshold, we tune the value of $K$ via grid search for each model-dataset pair. As such, $K$ was set to $0.125$,  
% $0.5$ and $1.0$, for IMDN, 
% TPSR and MobiSR-RCAN, respectively.

\begin{table}[t]
    \centering
    % \vspace{-0.2cm}
    \caption{\small QuantSR-WLopt vs. Heuristic Optimizers}
    % \vspace{-0.2cm}
    \resizebox{0.99\linewidth}{!}{
    \begin{threeparttable}
    \begin{tabular}{l l l c r l l l}
        \toprule
        \textbf{Model} & \textbf{Layers} & \textbf{Dataset} & \textbf{Target PSNR} & \textbf{Search} & \multicolumn{3}{c}{\textbf{BOPs Reduction}} \\
        \cline{6-8} 
        & & & \textbf{Drop} & \textbf{Time} & \textbf{WLopt} & \textbf{SA} & \textbf{GA} 
        % & \textbf{RS} 
        \\
        \midrule
        TPSR & 33 & B100 & 0.1 dB & 2.1 min & 1.96$\times$ & 1.68$\times$ & 1.59$\times$ 
        % & \textbf{TOADD} 
        \\
        TPSR & 33 & Urban100 & 0.1 dB & 9.6 min & 1.83$\times$ & 1.37$\times$ & 1.69$\times$ 
        % & \textbf{TOADD} 
        \\
        IMDN & 85 & B100 & 0.1 dB & 9.4 min & 1.93$\times$ & 1.66$\times$ & 1.44$\times$ 
        % & \textbf{TOADD} 
        \\
        IMDN & 85 & Urban100 & 0.1 dB & 22 min & 1.93$\times$ & 1.67$\times$ & 1.57$\times$ 
        % & \textbf{TOADD} 
        \\
        MobiSR-RCAN & 255 & B100 & 0.1 dB & 72 min & 2.00$\times$ & 1.72$\times$ & 1.56$\times$ 
        % & \textbf{TOADD} 
        \\
        MobiSR-RCAN & 255 & Urban100 & 0.1 dB & 177 min & 2.00$\times$ & 1.49$\times$ & 1.50$\times$ 
        % & \textbf{TOADD}
        \\
        \bottomrule
    \end{tabular}
    \begin{tablenotes}
        \item \small
        *SA setup: init. temperature $t_{0}$$=$$1$, cooling schedule at iter $i$: $t_i$$=$$t_{0} e^{-0.05i}$
        \item \phantom{*}GA setup: population size=$\lceil 0.25 \cdot \text{\#Layers} \rceil$ 
    \end{tablenotes}
    \end{threeparttable}
    }
    \label{tab:eval_wlopt}
\end{table}

% \vspace{-0.2cm}
\vspace{-0.5em}
\subsection{Evaluation of Wordlength Optimizer}
\label{sec:eval_wlopt}
% \vspace{-0.1cm}

We compare QuantSR-WLopt with three heuristic optimizers: \textit{1)}~simulated annealing (SA)~\cite{simulated_anneal1983science}, \textit{2)}~genetic algorithm (GA)~\cite{heuristics_combinat_opt1993book} and \textit{3)}~random search (RS). We compare the achieved BOPs reduction with respect to A16W8 given a PSNR drop constraint of 0.1~dB \textit{under the same search time budget}, across the evaluated SR DNNs and datasets B100 and Urban100. We utilize the runtime of QuantSR-WLopt as the search time budget and run each of the baselines 10 times on an Nvidia GTX1080Ti GPU, reporting the average best result in Table~\ref{tab:eval_wlopt}. 
First, as the attainable BOPs reduction over A16W8 is bounded to a maximum of 2$\times$, corresponding to INT8, we observe that our achieved reductions are very close to the peak performance, leaving little room for further improvement.
Furthermore, QuantSR-WLopt consistently outperforms all baselines, yielding a BOPs gain between 16\%-33\% (21.8$\%$ geo. mean) over SA and 8\%-34\% (24.7\% geo. mean) over GA. 
Finally, 
RS yielded designs that violated the PSNR constraint in the vast majority of runs and hence we omit it from Table~\ref{tab:eval_wlopt}. 

All three baseline optimizers are iterative and can quickly determine the next candidate design point to evaluate. As such, these strategies would be suitable in cases where the objective function (BOPs and PSNR in our setting) is cheap to evaluate. Nevertheless, as PSNR is costly to evaluate and the design space is combinatorially large, the more structured search approach of our QuantSR-WLopt is more effective in yielding a hybrid-precision design that lies close to the theoretical maximum of 2$\times$ BOPs reduction.

\setlength{\tabcolsep}{2pt}
\begin{table}[t]
    \centering
    % \vspace{-0.2cm}
    \caption{\small Quality Comparison with Baselines ($\times$4 Upscaling)}
    % \vspace{-0.2cm}
    \resizebox{0.9\linewidth}{!}{
    \begin{threeparttable}
        \begin{tabular}{l || llll }
            \toprule
            \textbf{Model} & \multicolumn{4}{c}{\textbf{Average PSNR/SSIM}}  \\
            \textbf{Variant} & \multicolumn{1}{c}{Set5} & \multicolumn{1}{c}{Set14} & \multicolumn{1}{c}{B100} & \multicolumn{1}{c}{Urban100} \\
            
            \midrule
    
            \multicolumn{1}{c}{} & \multicolumn{4}{c}{\textbf{TPSR} - (Depth=33, Params=61K)}  \\
            \hline
            FP32 & 31.10/0.8779 & 27.95/0.7663 & \textbf{27.15}/0.7214 & \textbf{24.97}/0.7456  \\
            
            FP32W8 & \textbf{30.92}/0.8737 & \textbf{27.85}/0.7634 &  27.08/0.7190 & 24.90/0.7423  \\
            
            FP16 & 31.10/0.8779 &  27.95/0.7663 & 27.15/0.7214 & 24.97/0.7456 \\
            
            INT8 & 30.75/0.8669 & 27.74/0.7573 & 26.99/0.7136 & 24.82/0.7362 \\
            
            A16W8 & 30.91/0.8736 & 27.83/0.7630 & 27.07/0.7189 & 24.88/0.7417 \\
            
            \hline
            % \tool w/o DRE & 30.89/0.8725 & 27.81/0.7614 & 27.04/0.7166 & 24.87/0.7407 \\
            
            % \tool w RandDRE & 30.89/0.8726 & 27.81/0.7615 & 27.05/0.7169 & 24.87/0.7408 \\
            
            \tool & 30.91/0.8730 & 27.83/0.7620 & 27.05/0.7170 & 24.88/0.7411 \\
            
            \hline 
            
            \multicolumn{1}{c}{} & \multicolumn{4}{c}{\textbf{IMDN} - (Depth=85, Params=698K)} \\
            \hline
            FP32 &  32.21/0.8948 & 28.58/0.7811 & \textbf{27.55}/0.7351 & 26.04/0.7837 \\
            
            FP32W8 & \textbf{32.04}/0.8921 & \textbf{28.46}/0.7795 & 27.47/0.7338 &  \textbf{25.92}/0.7814 \\
            
            FP16 & 32.21/0.8948 & 28.56/0.7809 & 27.52/0.7333 & 26.04/0.7837 \\
            
            INT8 & 31.86/0.8865 & 28.31/0.7749 & 27.35/0.7295 & 25.80/0.7753 \\
            
            A16W8 & 31.96/0.8913 & 28.38/0.7788 & 27.41/0.7336 & 25.85/0.7795 \\
            
            \hline
            % \tool w/o DRE & 31.94/0.8900 & 28.36/0.7775 & 27.38/0.7317 & 25.83/0.7776 \\
            
            % \tool w RandDRE & 31.97/0.8903 & 28.38/0.7773 & 27.40/0.7318 & 25.83/0.7774 \\
            
            \tool & 32.01/0.8911 & 28.47/0.7781 & 27.45/0.7325 & 25.89/0.7787 \\
            
            \hline 
            \multicolumn{1}{c}{} & 
            \multicolumn{4}{c}{\textbf{MobiSR-RCAN} - (Depth=255, Params=148K)} \\
            \hline
            FP32 &  \textbf{31.73}/0.8873 & 28.23/0.7729 & \textbf{27.33}/0.7283 & \textbf{25.34}/0.7615 \\
            
            FP32W8 & 31.71/0.8865 & \textbf{27.82}/0.7726 & 27.31/0.7282 &  25.33/0.7611 \\
            
            FP16 & 31.73/0.8873 & 28.23/0.7729 & 27.32/0.7283 & 25.34/0.7615 \\
            
            INT8 & 31.03/0.8793 & 27.76/0.7651 & 27.02/0.7225 & 24.97/0.7499 \\
            
            A16W8 & 31.10/0.8813 & 27.80/0.7668 & 27.06/0.7244 & 24.99/0.7517 \\
            
            \hline
            % \tool w/o DRE & 31.07/0.8803 & 27.76/0.7652 & 27.03/0.7227 & 24.97/0.7499 \\
            
            % \tool w RandDRE & 31.18/0.8811 & 27.86/0.7663 & 27.09/0.7234 & 25.12/0.7537 \\
            
            \tool & 31.69/0.8851 &  28.14/0.7696 & 27.27/0.7255 & 25.24/0.7557 \\
            
            \bottomrule
        \end{tabular}
    % }
        \begin{tablenotes}
            \item \small
            *Bold indicates the designs whose quality defines \tool's PSNR drop constraint.
        \end{tablenotes}
    % \vspace{0.2cm}
    % \vspace{-0.2cm}
    \end{threeparttable}
    \label{tab:psnr}
    }
\end{table}

% \vspace{-0.3cm}
\vspace{-0.5em}
\subsection{Evaluation of Neural Image Codec}
\label{sec:eval_neural_img_codec}
% \vspace{-0.1cm}

\textbf{Runtime Overheads.}
To evaluate the overhead of estimating new scale factors and zero points for each of the selected DRE layers, we measured the inference time, across 50 inferences, for each of the models with and without DRE enabled for these layers. 
Overall, across all DNNs, the average time overhead of DRE was $4.26\%$ (up to $6.40\%$) and $1.53\%$ (up to $4.58\%$) for B100 and Urban100, respectively.

Another potential overhead introduced by \tool is the cost of switching between partitions with distinct bitwidths (\textit{i.e.}~INT8 vs. A16W8).
To evaluate this, we measured the switching times across 50 inferences for each of the DNNs, using the partitions selected by \tool.
The average partition-switching overhead over the inference time across DNNs was $0.34\%$ (up to $0.84\%$) and $1.04\%$ (up to $2.41\%$), for B100 and Urban100, respectively, with an average latency overhead of $39.25\mu{s}$ (up to $53\mu{s}$) per partition.

\setlength{\tabcolsep}{3pt}
\begin{table}[t]
    \centering
    % \vspace{-0.2cm}
    \caption{\small \red{Ablation Study on LRA-based layer selection and using DRE altogether.}}
    % \vspace{-0.2cm}
    \resizebox{0.9\linewidth}{!}{
    \begin{threeparttable}
        \begin{tabular}{l || c c || llll }
            \toprule
            \textbf{Model} & & & \multicolumn{4}{c}{\textbf{Average PSNR/SSIM}}  \\
            \textbf{Variant} & \textbf{LRA} & \textbf{DRE} & \multicolumn{1}{c}{Set5} & \multicolumn{1}{c}{Set14} & \multicolumn{1}{c}{B100} & \multicolumn{1}{c}{Urban100} \\
            
            \midrule
    
            \multicolumn{1}{c}{} & & & \multicolumn{4}{c}{\textbf{TPSR} - (Depth=33, Params=61K)}  \\         
            \hline
            w/o DRE & \xmark & \xmark & 30.89/0.8725 & 27.81/0.7614 & 27.04/0.7166 & 24.87/0.7407 \\
            
            RandDRE & \xmark & \cmark & 30.89/0.8726 & 27.81/0.7615 & 27.05/0.7169 & 24.87/0.7408 \\
            
            \tool & \cmark & \cmark & 30.91/0.8730 & 27.83/0.7620 & 27.05/0.7170 & 24.88/0.7411 \\
            
            \hline 
            
            \multicolumn{1}{c}{} & & & \multicolumn{4}{c}{\textbf{IMDN} - (Depth=85, Params=698K)} \\
            \hline
            w/o DRE & \xmark & \xmark & 31.94/0.8900 & 28.36/0.7775 & 27.38/0.7317 & 25.83/0.7776 \\
            
            RandDRE & \xmark & \cmark & 31.97/0.8903 & 28.38/0.7773 & 27.40/0.7318 & 25.83/0.7774 \\
            
            \tool & \cmark & \cmark & 32.01/0.8911 & 28.47/0.7781 & 27.45/0.7325 & 25.89/0.7787 \\
            
            \hline 
            \multicolumn{1}{c}{} & & & \multicolumn{4}{c}{\textbf{MobiSR-RCAN} - (Depth=255, Params=148K)} \\
            
            \hline
            w/o DRE & \xmark & \xmark & 31.07/0.8803 & 27.76/0.7652 & 27.03/0.7227 & 24.97/0.7499 \\
            
            RandDRE & \xmark & \cmark & 31.18/0.8811 & 27.86/0.7663 & 27.09/0.7234 & 25.12/0.7537 \\
            
            \tool & \cmark & \cmark & 31.69/0.8851 &  28.14/0.7696 & 27.27/0.7255 & 25.24/0.7557 \\
            
            \bottomrule
        \end{tabular}
    % }
        \begin{tablenotes}
            \item \small
            % *Bold indicates the designs whose quality defines \tool's PSNR drop constraint.
        \end{tablenotes}
    % \vspace{0.2cm}
    % \vspace{-0.2cm}
    \end{threeparttable}
    \label{tab:ablation}
    }
\end{table}

\textbf{\red{Ablation Study of LRA and DRE.}}
\red{We conduct an ablation study on \textit{i)}~our LRA-based layer selection and \textit{ii)}~using DRE altogether, in order to disentangle their impact on the achieved quality. For each model in Table~\ref{tab:ablation}, we show the achieved PSNR/SSIM for the following configurations: \textit{i)}~w/o DRE, where we use \tool's selected bitwidths for each layer, but the scale factors and zero points of each activations tensor are derived \textit{a priori} based on the maximum range encountered in the calibration set and remain fixed during deployment; \textit{ii)}~RandDRE, where we use \textit{1)}~\tool's selected bitwidths for each layer, and uniform probability to \textit{2)}~randomly select the number of DRE layers and then \textit{3)}~randomly select the layers. For RandDRE, we report the average quality across 10 runs; and \textit{iii)}~NAWQ-SR, our method that selectively applies DRE using our LRA-based layer selection scheme.}

\red{Across all models and datasets, we observe that although RandDRE already provides quality gains over w/o DRE, the informed layer selection of the complete \tool contributes significant additional gains. Specifically, DRE with LRA yield similar or higher quality, with gains of up to 0.02 dB (0.015 dB average) for TPSR, 0.11 dB (0.08 dB average) for IMDN and 0.62 dB (0.38 dB average) for MobiSR-RCAN. 
Notably, the gains of DRE are higher for deeper models as these models are more affected by the accumulation of quantization errors across layers, resulting in a larger drop in visual quality. As DRE significantly reduces the degree of error accumulation, it results in significant qualitative improvements in IMDN and MobiSR-RCAN: specifically, the mitigation of undesirable quantization artifacts on both texture and color as shown in Fig.~\ref{fig:qualitative}.}

\red{From a computational perspective, RandRE often picks a suboptimal set of layers, resulting in 19.5$\times$ average higher overhead compared to \tool's DRE layer selection. Instead, our LRA-based approach offers the advantage of determining in a single step both the number and the DRE layers that have the highest impact on quality. As a result, although a naive application of DRE can still yield a performance improvement, our more selective layer selection method achieves a better trade-off that combines both higher quality and lower latency overhead, and is, thus, an essential component of the proposed system.}

\red{Overall, as shown in Fig.~\ref{fig:speedup} and Table~\ref{tab:psnr}, the Neural Image Codec presents a very reasonable overhead considering its latency and visual quality when compared to the full- (FP32) and lowest-precision (INT8) baselines.}

% \vspace{-0.3cm}
% \vspace{-0.4em}
\subsection{Comparison with Highly Optimized Baselines}
\label{sec:eval_baselines}
% \vspace{-0.1cm}

\begin{table}[t]
    \centering
    % \vspace{-0.4cm}
    \caption{\small Speedup over Highly Optimized Baselines}
    % \vspace{-0.2cm}
    \resizebox{0.9\linewidth}{!}{
    \begin{threeparttable}
    \begin{tabular}{l l c c}
        \toprule
        \textbf{Model} & \textbf{Baseline} & \textbf{Speedup on SDM865} & \textbf{\red{Speedup on SDM888}} 
        \\
        & & \textbf{B100/Urban100} & \textbf{\red{B100/Urban100}} 
        \\
        \midrule
         & FP32-CPU & 40.89$\times$/40.80$\times$ & 55.55$\times$/55.44$\times$
        \\
        TPSR & FP16-GPU & 12.54$\times$/12.51$\times$ & 16.70$\times$/16.66$\times$
        \\
         & A16W8-NPU & \phantom{1}6.08$\times$/\phantom{1}6.07$\times$ & \phantom{1}8.08$\times$/\phantom{1}8.06$\times$
        \\
         & INT8-NPU & \phantom{1}3.65$\times$/\phantom{1}3.64$\times$ & \phantom{1}4.61$\times$/\phantom{1}4.60$\times$
        \\
        \midrule
         & FP32-CPU & \phantom{1}9.97$\times$/\phantom{1}9.91$\times$ & 13.69$\times$/13.62$\times$
        \\
        IMDN & FP16-GPU & \phantom{1}1.88$\times$/\phantom{1}1.87$\times$ & \phantom{1}2.46$\times$/\phantom{1}2.44$\times$
        \\
         & A16W8-NPU & \phantom{1}1.89$\times$/\phantom{1}1.88$\times$ & \phantom{1}1.83$\times$/\phantom{1}1.82$\times$
        \\
         & INT8-NPU  & \phantom{1}1.59$\times$/\phantom{1}1.58$\times$ & \phantom{1}1.52$\times$/\phantom{1}1.51$\times$
        \\
        \midrule
         & FP32-CPU & 26.11$\times$/26.47$\times$ & 34.59$\times$/35.02$\times$
        \\
        MOBISR-RCAN & FP16-GPU & \phantom{1}7.04$\times$/\phantom{1}7.14$\times$ & 9.33$\times$/9.45$\times$ 
        \\
         & A16W8-NPU & \phantom{1}3.87$\times$/\phantom{1}3.92$\times$ & \phantom{1}4.80$\times$/\phantom{1}4.86$\times$ 
        \\
         & INT8-NPU & \phantom{1}2.04$\times$/\phantom{1}2.07$\times$ & \phantom{1}2.49$\times$/\phantom{1}2.52$\times$
        \\
        \midrule
        \midrule
        & FP32-CPU & 25.69$\times$ (22.02$\times$) & 34.65$\times$ (29.77$\times$) \\
        {Average (geo. mean)} & FP16-GPU & \phantom{1}7.16$\times$ (\phantom{1}5.51$\times$) & \phantom{1}9.51$\times$ (\phantom{1}7.27$\times$) \\
        & A16W8-NPU & \phantom{1}3.95$\times$ (\phantom{1}3.55$\times$) & \phantom{1}4.91$\times$ (\phantom{1}4.14$\times$) \\
        & INT8-NPU & \phantom{1}2.43$\times$ (\phantom{1}2.28$\times$) & \phantom{1}2.87$\times$ (\phantom{1}2.59$\times$) \\
        
        \bottomrule
    \end{tabular}
    \end{threeparttable}
    }
    \label{tab:speedup_baselines}
\end{table}

\begin{figure}[t]
    \centering
    % \vspace{-0.3cm}
    \begin{subfigure}{0.235\textwidth}
        \centering
        % \fbox{
        \includegraphics[trim={0.5cm 0.4cm 1cm 1cm},clip,width=1\columnwidth]{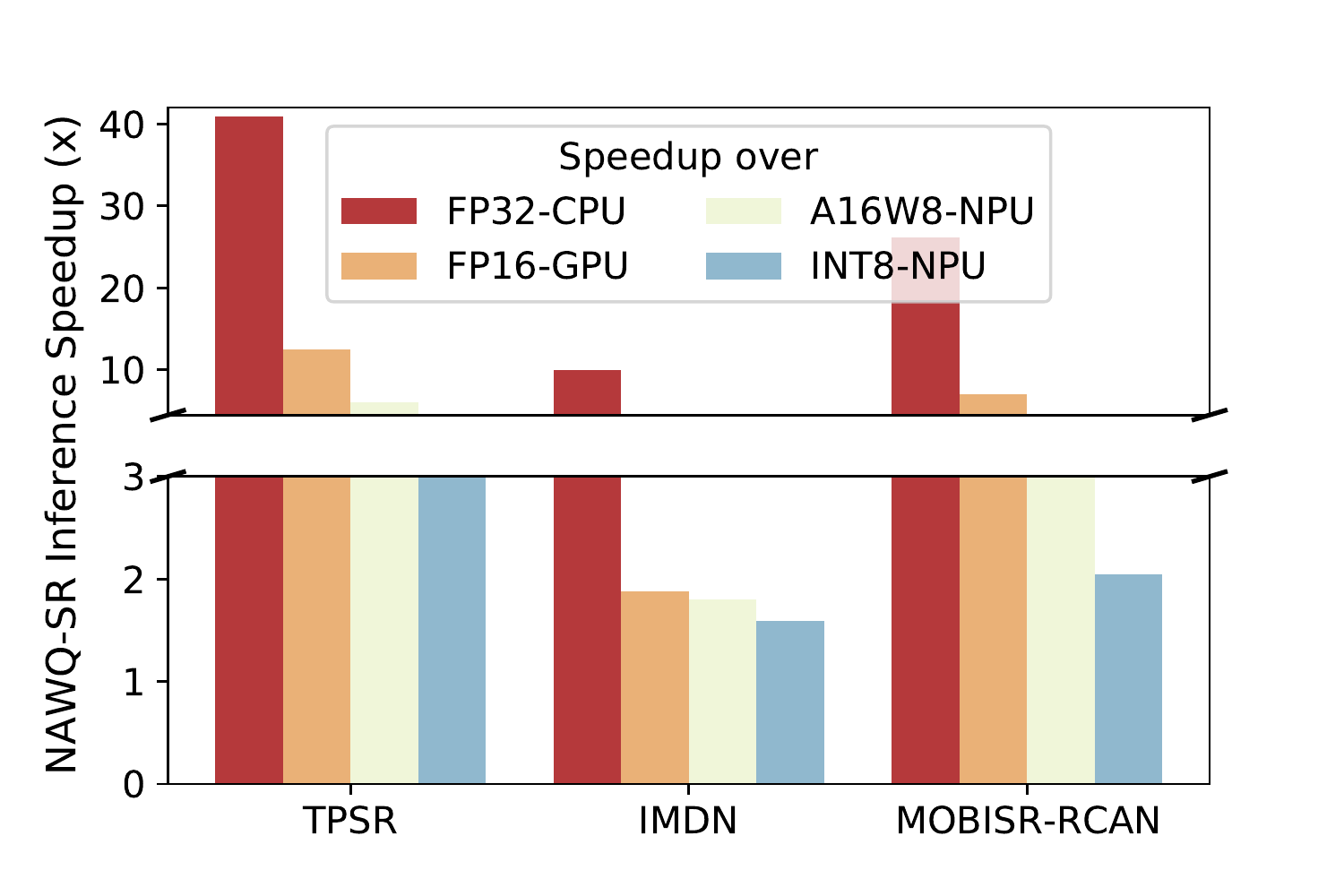}
        % }
        \vspace{-0.6cm}
        \caption{B100 on SDM865.}
        \label{fig:speedup_b100}
    \end{subfigure}
    \begin{subfigure}{0.235\textwidth}
        \centering
        \includegraphics[trim={0.5cm 0.4cm 1cm 1.0cm},clip,width=1\columnwidth]{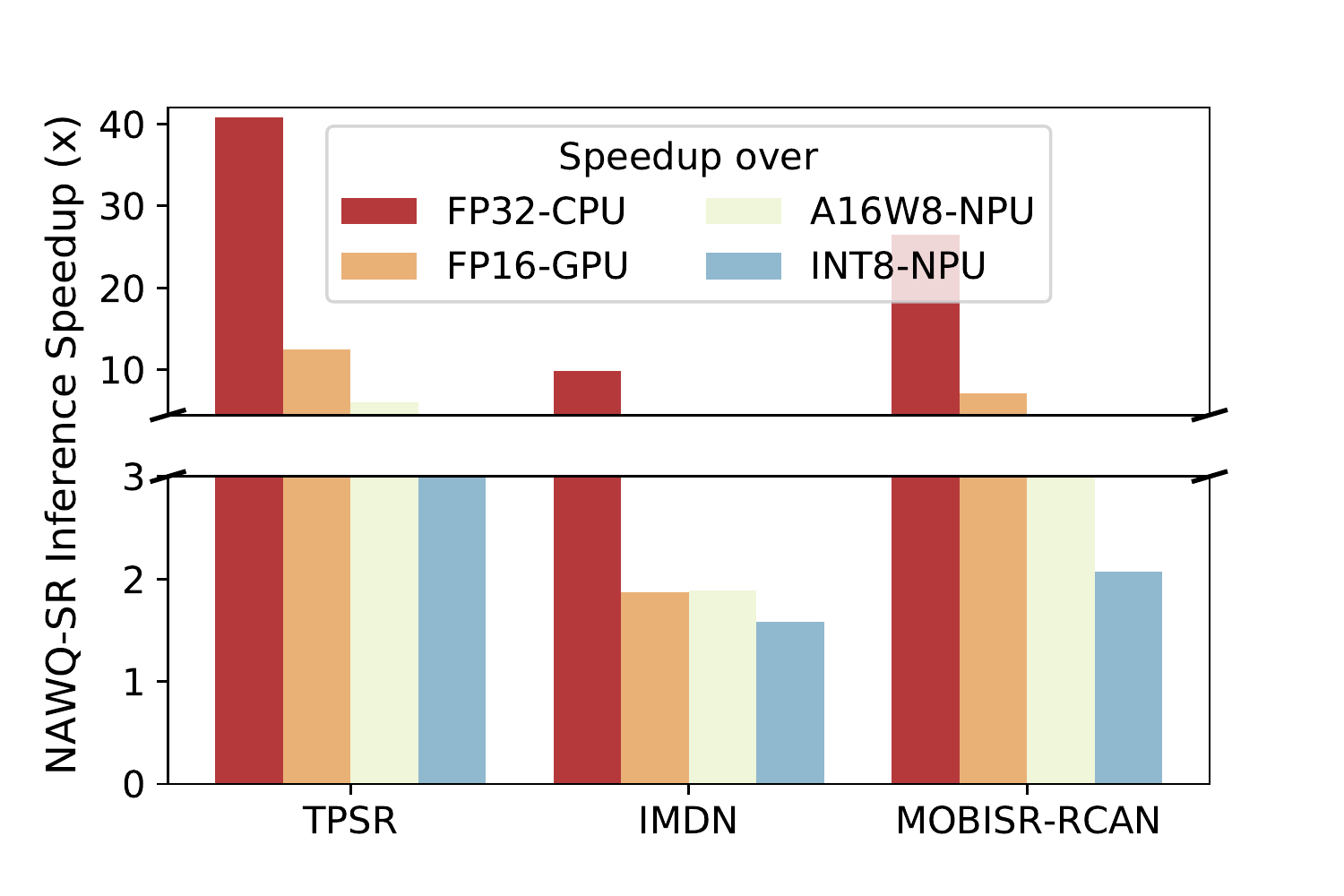}
        \vspace{-0.6cm}
        \caption{Urban100 on SDM865.}
        \label{fig:speedup_urban100}
    \end{subfigure}

    \begin{subfigure}{0.235\textwidth}
        \centering
        % \fbox{
        \includegraphics[trim={0.5cm 0.4cm 1cm 1cm},clip,width=1\columnwidth]{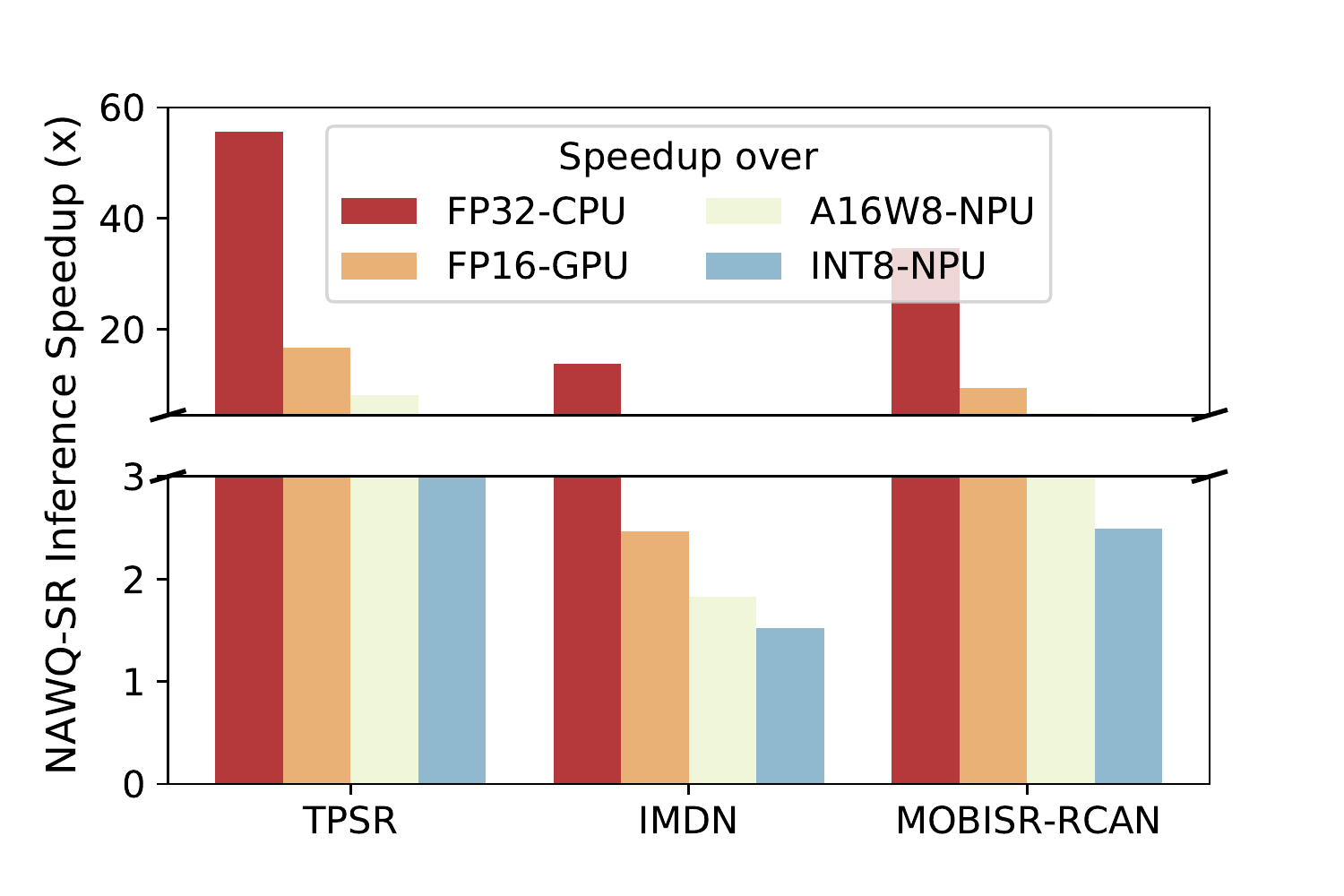}
        % }
        \vspace{-0.6cm}
        \caption{B100 on SDM888.}
        \label{fig:speedup_b100}
    \end{subfigure}
    \begin{subfigure}{0.235\textwidth}
        \centering
        \includegraphics[trim={0.5cm 0.4cm 1cm 1.0cm},clip,width=1\columnwidth]{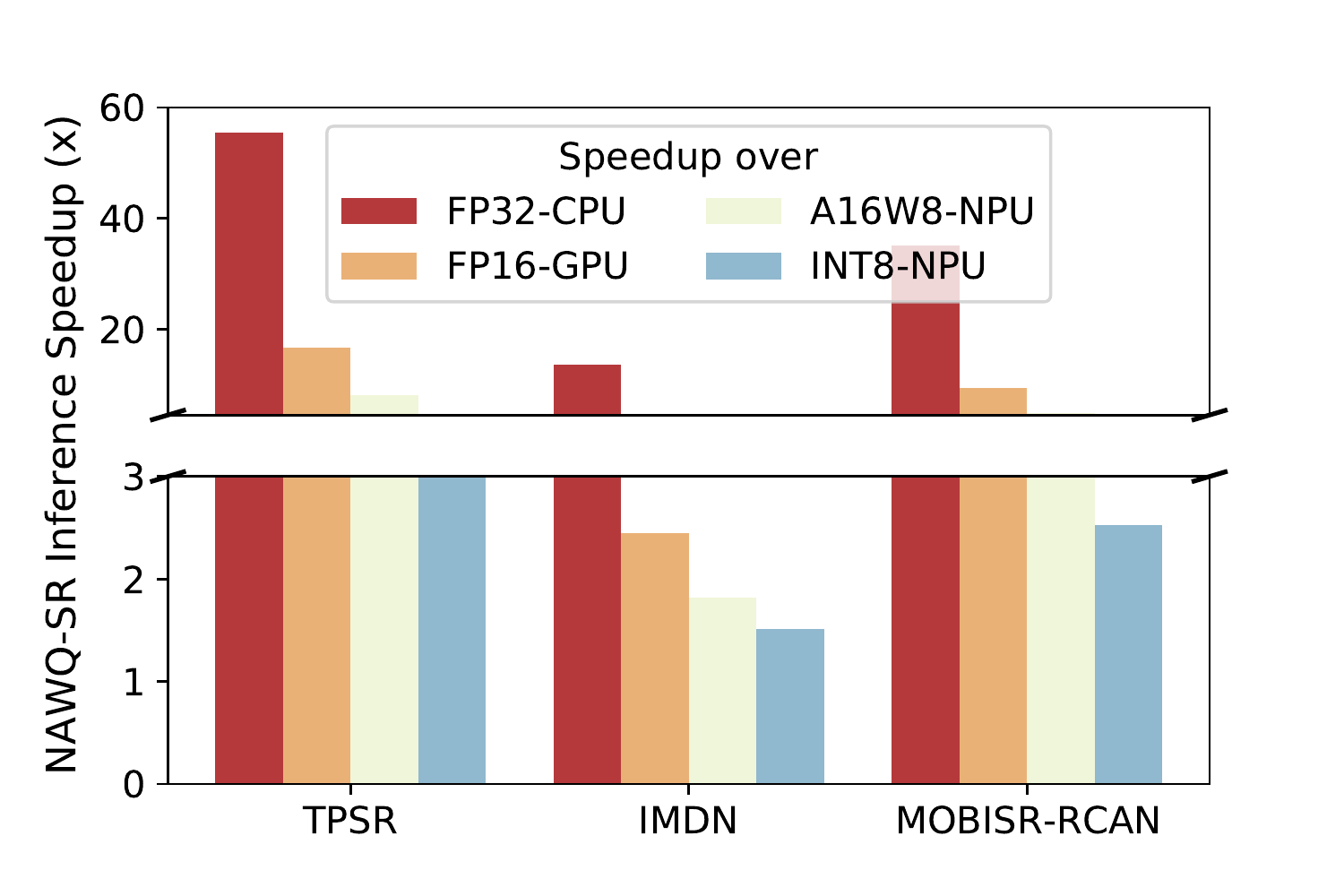}
        \vspace{-0.6cm}
        \caption{Urban100 on SDM888.}
        \label{fig:speedup_urban100}
    \end{subfigure}
    
    % \vspace{-0.9em}
    \caption{\tool's inference speedup over highly optimized baselines across SR DNNs, \red{targeting SDM865 and SDM888}.}
    \label{fig:speedup}
\end{figure}

This section presents a comparison of \tool with the following: FP32-CPU, FP16-GPU, INT8-NPU and A16W8-NPU designs, obtained through SNPE. These represent status-quo implementations that have been highly optimized using the SNPE compiler targeting each of the available processors. In the case of INT8-NPU, we allow the layers to be executed on both HVX and HTA to obtain the fastest execution. Table~\ref{tab:psnr} presents the achieved quality and Fig.~\ref{fig:speedup} and Table~\ref{tab:speedup_baselines} depict the achieved speedup measured on SDM865 and SDM888 across models and datasets. 
We also report the quality after quantizing only the weights (FP32W8).

\textbf{Comparison to CPU/GPU Designs.}
With respect to the floating-point designs (FP32/FP16), \tool delivers quality within 0.1 dB of the original model's for the vast majority of cases. In cases where weights quantization has a significant impact on quality, \textit{e.g.}~FP32W8 leads to $\geq$0.1~dB drop over FP32 for Set5, Set14 and Urban100 in IMDN, our framework was optimized with a 0.1~dB tolerance with respect to FP32W8. This is achieved across all cases.
With respect to latency, \tool outperforms both CPU and GPU designs by up to 40.8$\times$ (22$\times$ geo. mean across models and datasets) and 12.5$\times$ (5.5$\times$ geo. mean), respectively, on SDM865 \red{and by up to 55.5$\times$ (29.7$\times$ geo. mean) and 13.6$\times$ (7.2$\times$ geo. mean), respectively, on SDM888}.

\textbf{Comparison to NPU Designs.}
With respect to A16W8-NPU, \tool outperforms its PSNR for IMDN and MobiSR-RCAN with an average gain of 0.05~dB for IMDN and 0.35~dB for MobiSR-RCAN across datasets. For TPSR, \tool generates mappings that either have slightly lower PSNR but still lie within the PSNR constraint with respect to FP32 (see B100), or meet the PSNR of A16W8-NPU. On the latency front, \tool provides up to 6$\times$ \red{and 8$\times$} faster execution than A16W8-NPU on SDM865 \red{and SDM888, respectively}, with a geometric mean of 3.55$\times$ \red{and 4.14$\times$ on the respective device} across models and datasets. Compared to INT8-NPU, \tool yields higher PSNR with an average of 0.09 dB for TPSR, 0.12 dB for IMDN and 0.39 dB for MobiSR-RCAN across the datasets.
\blue{With respect to latency, our system achieves up to 3.65$\times$ \red{and 4.61$\times$} faster processing than INT8-NPU on SDM854 \red{and SDM888, respectively}, with a geometric mean of 2.28$\times$ \red{and $2.59\times$ on each device} across models and datasets.}
\tool's speedup is attributed to our highly optimized memory-aware mapping of the pixel-shuffle upsampling layers (\S~\ref{sec:neural_codec}) which enables the uninterrupted execution of the SR DNNs on the NPU, without falling back to CPU or GPU.

Overall, the results demonstrate how the hybrid-precision approach and the better utilization of the NPU's capabilities provided by our system allow us to closely track the quality of floating-point execution, outperform current INT8 designs, while pushing beyond A16W8's quality in several cases.

% \vspace{-0.3cm}
\vspace{-0.5em}
\subsection{Comparison with the state-of-the-art \\ On-Device SR Systems}
\label{sec:baselines}
% \vspace{-0.1cm}

Here, we show the performance gains of \tool \blue{as a standalone framework} over the state-of-the-art on-device SR systems, MobiSR~\cite{mobisr2019mobicom} and SplitSR~\cite{splitsr2021imwut}\blue{, and the winning model of the 2021 MAI challenge~\cite{mobile_sr_challenge2021cvprw} on quantized SR on mobile NPUs, XLSR~\cite{xlsr2021cvprw} } (Table~\ref{tab:eval_sota}).
\blue{MobiSR, SplitSR and XLSR constitute the state-of-the-art image SR systems using heterogeneous processors, CPU, and NPU, respectively.} 
\blue{For fair comparisons, we reimplemented and ran MobiSR on the same device (SDM865).}
Both systems base their design on lightweight variants of RCAN~\cite{RCAN}.
% We also compare with the winning model of the 2021 MAI challenge~\cite{mobile_sr_challenge2021cvprw} on quantized SR on mobile NPUs, XLSR~\cite{xlsr2021cvprw}. 

% \textbf{Comparison with MobiSR.}
% MobiSR employs two models that are parallelized across the heterogeneous processors of the target device. The computationally heavier model is run on the CPU and GPU and the lightweight one on the NPU. MobiSR's scheduler divides the input image into patches and feeds them to each model-processor pair based on their difficulty; more difficult-to-upscale patches are sent for rapid processing to the NPU and easier patches are directed to the CPU and GPU in a load-balancing manner.
% In \cite{mobisr2019mobicom}, three system configurations are presented, each optimized for a different objective: \textbf{MobiSR-accuracy}, \textbf{-balanced} and \textbf{-latency} (Appendix \ref{app:mobisr}).
\textbf{Comparison with MobiSR.}
MobiSR employs two models that are parallelized across the heterogeneous processors of the target device. 
The computationally heavier model is run on the CPU and GPU and the lightweight one on the NPU. 
MobiSR's scheduler divides the input image into patches and feeds them to each model-processor pair based on their difficulty; more difficult-to-upscale patches are sent for rapid processing to the NPU and easier patches are directed to the CPU and GPU in a load-balancing manner.
Lee~\textit{et al.}~\cite{mobisr2019mobicom} present three system configurations, each optimized for a different objective:

\begin{itemize}[leftmargin=*,noitemsep,topsep=0pt]
    \item \textbf{MobiSR-accuracy}: The accuracy-optimized model pair, denoted by $(m_{\text{ref}}+m_{\text{clc}})$ in \cite{mobisr2019mobicom}. $m_{\text{ref}}$ denotes the original MobiSR-RCAN architecture. $m_{\text{clc}}$ employs group convolutions and channel-shuffle layers~\cite{shufflenetv1_2018cvpr,clcnet2018cvpr} to reduce the computational complexity of the original MobiSR-RCAN.
    
    \item \textbf{MobiSR-balanced}: The accuracy-latency balanced model pair, denoted by $(m_{\text{ref}}+m_{\text{s2}})$ in \cite{mobisr2019mobicom}. The compact model $m_{\text{s2}}$ goes beyond the channel shuffling of $m_{\text{clc}}$ and introduces channel splitting~\cite{shufflenetv2_2018eccv} and depthwise-separable convolutions~\cite{mobilenet2017arxiv} to further improve latency.
    
    \item \textbf{MobiSR-latency}: The latency-optimized model pair, denoted by $(m_{\text{clc}}+m_{\text{s2}})$ in \cite{mobisr2019mobicom}. This model pair combines the complexity-reduction techniques of the high-accuracy and balanced model pairs, delivering fast processing at the expense of degraded visual quality.
\end{itemize}
% , each optimized for a different objective:
% 
% \begin{itemize}[leftmargin=*,noitemsep,topsep=0pt]
%     \item \textbf{MobiSR-accuracy}: The accuracy-optimized model pair, denoted by $(m_{\text{ref}}+m_{\text{clc}})$ in \cite{mobisr2019mobicom}. $m_{\text{ref}}$ denotes the original MobiSR-RCAN architecture. $m_{\text{clc}}$ employs group convolutions and channel-shuffle layers~\cite{shufflenetv1_2018cvpr,clcnet2018cvpr} to reduce the computational complexity of the original MobiSR-RCAN.
    % 
%     \item \textbf{MobiSR-balanced}: The accuracy-latency balanced model pair, denoted by $(m_{\text{ref}}+m_{\text{s2}})$ in \cite{mobisr2019mobicom}. The compact model $m_{\text{s2}}$ goes beyond the channel shuffling of $m_{\text{clc}}$ and introduces channel splitting~\cite{shufflenetv2_2018eccv} and depthwise-separable convolutions~\cite{mobilenet2017arxiv} to further improve latency.
    % 
%     \item \textbf{MobiSR-latency}: The latency-optimized model pair, denoted by $(m_{\text{clc}}+m_{\text{s2}})$ in \cite{mobisr2019mobicom}. This model pair combines the complexity-reduction techniques of the high-accuracy and balanced model pairs, delivering fast processing at the expense of degraded visual quality.
% \end{itemize}
% 
% To perform a fair comparison, we tune the total-variation threshold of each MobiSR variant, so that it meets 0.1 dB PSNR drop with respect to the original MobiSR-RCAN (Appendix~\ref{app:tv_tuning}).
Furthermore, MobiSR introduces a parameter named total-variation (TV) threshold that tunes the accuracy-latency trade-off of each pair of models.
To perform a fair comparison, we tune the TV threshold of each MobiSR variant, so that it meets 0.1 dB PSNR drop with respect to the original MobiSR-RCAN.
As such, we set TV to {\small$\left<8,8,6,6\right>$} for Set5, Set14, B100 and Urban100 for MobiSR-accuracy, {\small$\left<8,8,6,8\right>$} for MobiSR-balanced and to 10 for all datasets for MobiSR-latency.
Accordingly, we apply \tool over MobiSR-RCAN with the same PSNR drop tolerance.
% 
% Furthermore, MobiSR introduces a parameter named total-variation (TV) threshold that controls the accuracy-latency trade-off of each pair of models.
% To perform a fair comparison, we tune the TV threshold of each MobiSR variant, so that it meets 0.1 dB PSNR drop with respect to the original MobiSR-RCAN (Appendix~\ref{app:mobisr}).
% As such, we set TV to {\small$\left<8,8,6,6\right>$} for Set5, Set14, B100 and Urban100 for MobiSR-accuracy, {\small$\left<8,8,6,8\right>$} for MobiSR-balanced and to 10 for all datasets for MobiSR-latency.
% Accordingly, we apply \tool over MobiSR-RCAN with the same PSNR drop tolerance.

\begin{table}[t]
    \centering
    % \vspace{-0.2cm}
    % \vspace{-0.4cm}
    \caption{\small Comparison with Existing On-Device SR Systems}
    % \vspace{-0.15cm}
    \resizebox{1.\linewidth}{!}{
        \begin{tabular}{l l c l l l l l l l l}
            \toprule
            \textbf{System} & \textbf{Model} & \textbf{Memory} & \multicolumn{4}{c}{\textbf{Average PSNR/SSIM}}  \\
            & & \textbf{(KB)} & \multicolumn{1}{c}{Set5} & \multicolumn{1}{c}{Set14} & \multicolumn{1}{c}{B100} & \multicolumn{1}{c}{Urban100} 
            \\
            \midrule
            Original & MobiSR-RCAN & 594 & 31.73/0.8873 & 28.23/0.7729 & 27.33/0.7283 & 25.34/0.7615 \\
            \hline
            MobiSR & (accuracy) & 623 & 31.37/0.8787 & 28.10/0.7707 & 27.28/0.7258 & 25.28/0.7591 \\
            MobiSR & (balanced) & 610 & 30.89/0.8590 & 27.98/0.7650 & 27.23/0.7207 & 25.31/0.7598 \\
            MobiSR & (latency) & 134 & 31.05/0.8762 & 27.87/0.7640 & 27.11/0.7208 & 24.85/0.7415 \\
            \tool & MobiSR-RCAN & 148 & 31.69/0.8851 & 28.14/0.7696 & 27.28/0.7261 & 25.25/0.7558 \\
            \midrule
            
            SplitSR & (accuracy) & 679 & 31.76/0.8982 & 28.29/0.7916 & 27.39/0.7491 & 25.46/0.7795 \\
            \tool & IMDN & 698 & 32.01/0.8911 & 28.47/0.7781 & 27.45/0.7325 & 25.89/0.7787 \\
            \midrule
            
            SplitSR & (latency) & 367 & 31.53/0.8950 & 28.18/0.7887 & 27.28/0.7458 & 25.20/0.7704 \\
            \tool & MobiSR-RCAN & 148 & 31.69/0.8851 & 28.14/0.7696 & 27.28/0.7261 & 25.25/0.7558 \\
            \bottomrule
        \end{tabular}
    }
    \resizebox{1.\linewidth}{!}{
        \begin{tabular}{l l l l l c c c c}
            \toprule
            \textbf{System} & \textbf{Model} & \textbf{Upscale} & \textbf{Precision} & \textbf{Memory} & \multicolumn{2}{c}{\textbf{Average PSNR/SSIM}} & \multicolumn{2}{c}{\textbf{Speedup}}  \\
            
            & & \textbf{Factor} & & \textbf{(KB)} & B100 & Urban100 & over CPU & over XLSR \\
            
            \midrule
            
            XLSR & - & $\times$3 & FP32 & 268 & \multicolumn{1}{l}{28.55/-} & \multicolumn{1}{l}{26.71/-} &  \phantom{2}1.00$\times$ & 1.00$\times$ \\
            
            XLSR & - & $\times$3 & INT8 & 67 & \multicolumn{1}{l}{28.05-28.35/-} & \multicolumn{1}{l}{26.21-26.51/-} &  28.50$\times$ & 1.00$\times$ \\
            
            \tool & TPSR & $\times$3 & hybrid & 61 & 28.56/0.7861 & 26.78/0.8126 & 54.41$\times$ & 1.91$\times$ \\
            
            \bottomrule
        \end{tabular}
    }
    \label{tab:eval_sota}
\end{table}

\begin{figure}[t]
    % \vspace{-0.5cm}
    \centering
    \subfloat[Comparison on B100.]{
        \centering
        \includegraphics[trim={0.5cm 0.0cm 1cm 1cm},clip,width=0.235\textwidth]{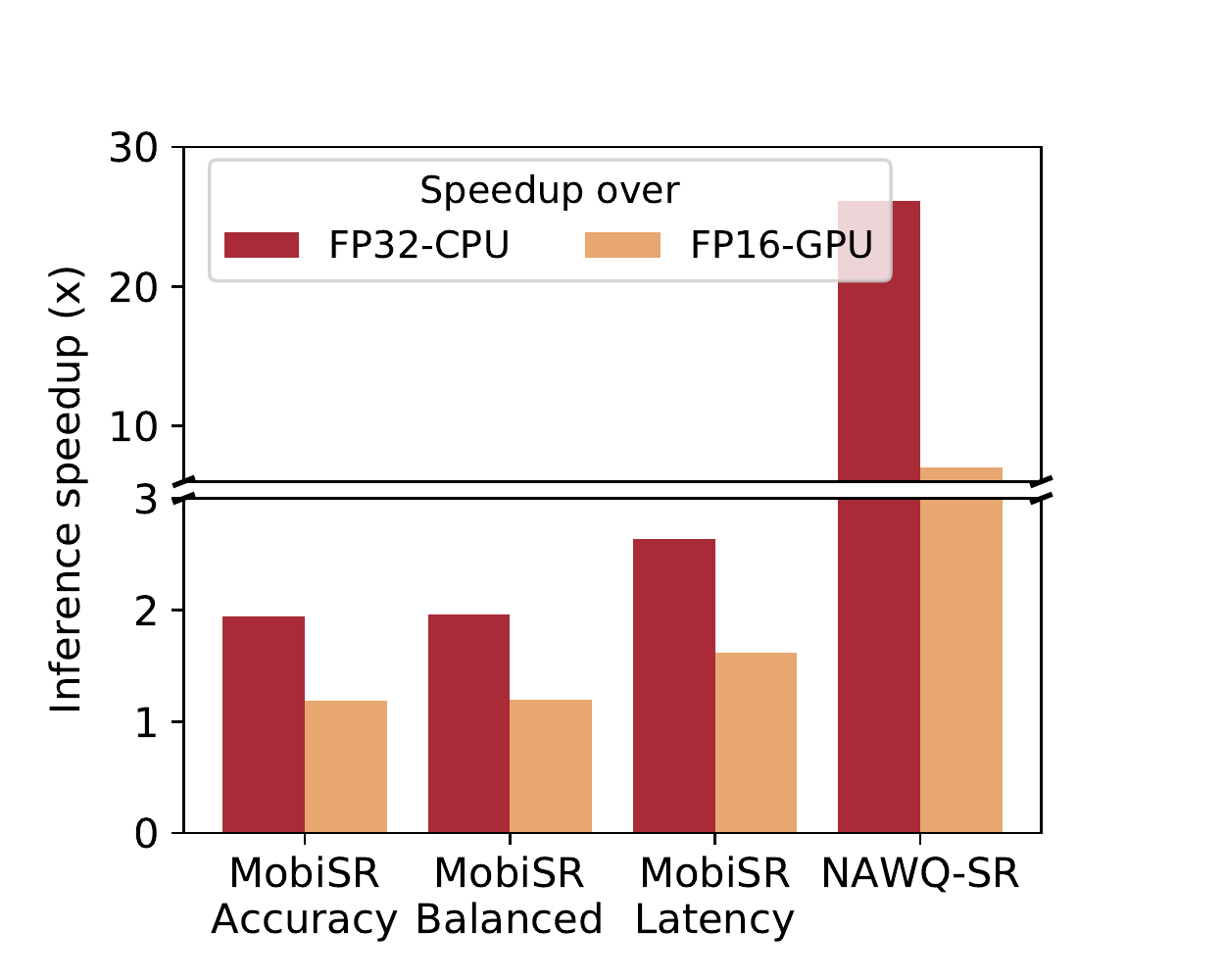}
        % \vspace{-0.6cm}
        % \caption{Comparison on B100.}
        \label{fig:mobisr_b100}
    }
    \subfloat[Comparison on Urban100.]{
        \centering
        \includegraphics[trim={0.5cm 0.0cm 1cm 1.0cm},clip,width=0.235\textwidth]{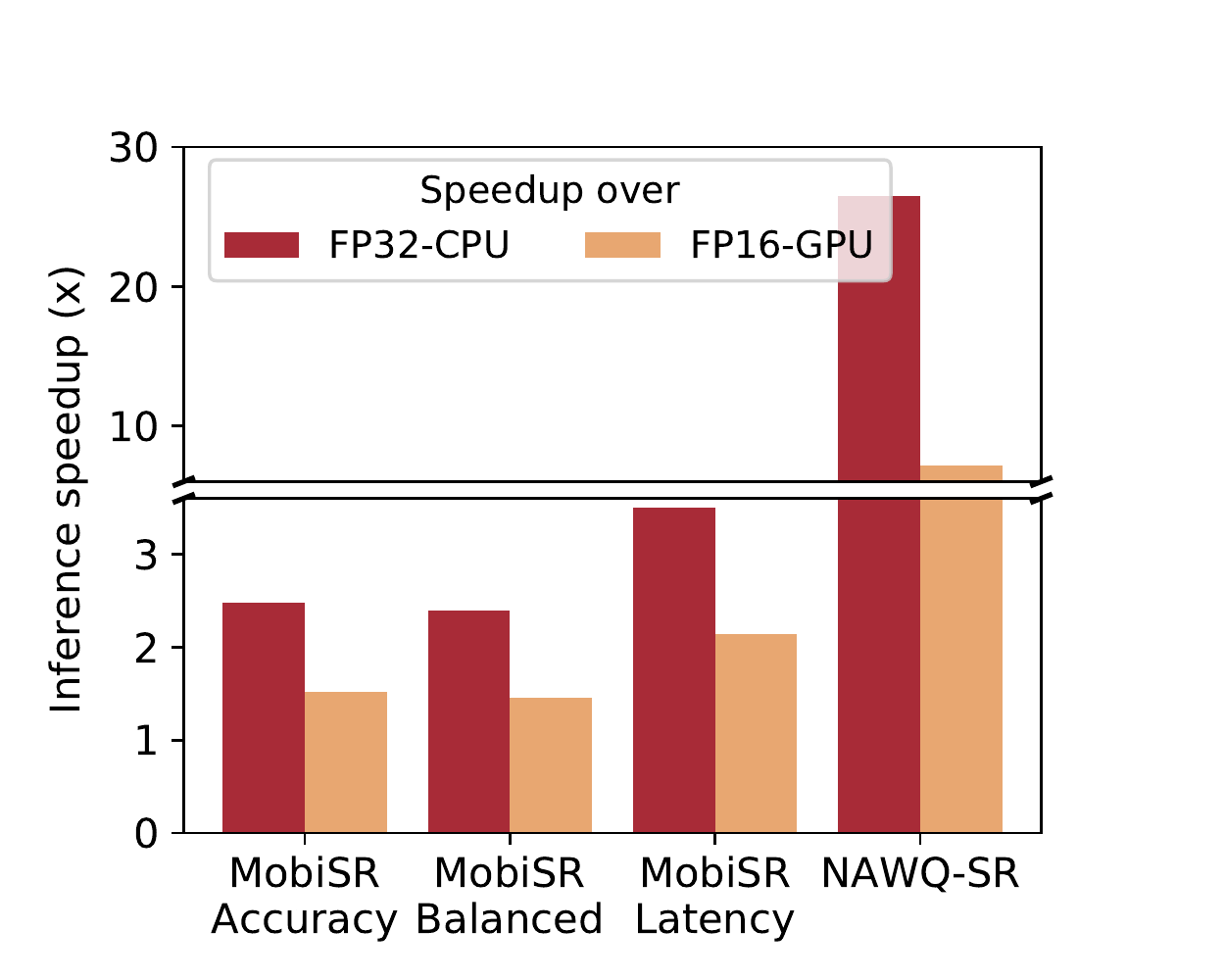}
        % \vspace{-0.6cm}
        % \caption{Comparison on Urban100.}
        \label{fig:mobisr_urban100}
    }
    \vspace{-0.2cm}
    \caption{Speedup comparison against MobiSR.}
    \vspace{0.1cm}
    \label{fig:mobisr}
\end{figure}

Fig.~\ref{fig:mobisr} depicts the actual speedup achieved by MobiSR and \tool over highly optimized CPU and GPU implementations on Urban100. 
% \tool outperforms MobiSR yielding up to 11.1$\times$ and 4.9$\times$ higher speedup over the CPU and GPU mapping, respectively.
% Similarly, on B100 (Appendix-Fig.~\ref{fig:app:mobisr_b100}), \tool achieves up to 13.4$\times$ and 5.9$\times$ higher speedup over MobiSR compared to the CPU and GPU implementations, respectively. 
% Similarly, on Urban100, \tool achieves up to 11.1$\times$ and 4.9$\times$ higher speedup over MobiSR compared to the CPU and GPU implementations, respectively. 
% Due to its approach of quantizing the compact DNN that runs on the NPU, MobiSR has to compensate for the PSNR drop by scheduling a significant portion of patches to the expensive CPU- and GPU-pinned model. Instead, through the combination of hybrid-precision execution and DRE, \tool alleviates the destructive effect of quantization on quality and enables the fast processing of all patches on the NPU. Overall, \tool achieves an average speedup improvement of 7.93$\times$ (7.17$\times$ geo. mean) across models and datasets.
On B100, \tool outperforms MobiSR yielding up to 13.4$\times$ and 5.9$\times$ higher speedup over the CPU and GPU mapping, respectively. 
Similarly, on Urban100, \tool achieves up to 11.1$\times$ and 4.9$\times$ higher speedup over MobiSR compared to the CPU and GPU implementations, respectively. Due to its approach of quantizing the compact DNN that runs on the NPU, MobiSR has to compensate for the PSNR drop by scheduling a significant portion of patches to the expensive CPU- and GPU-pinned model. Instead, through the combination of hybrid-precision execution and DRE, \tool alleviates the destructive effect of quantization on quality and enables the fast processing of all patches on the NPU. Overall, \tool achieves an average speedup improvement of 7.93$\times$ (7.17$\times$ geo. mean) across models and datasets.

\textbf{Comparison with SplitSR.}
% SplitSR introduces a compact residual block (SplitSRBlock) and modifies RCAN to allow for a configurable accuracy-computational cost trade-off, using a single model. Two system configurations were presented in \cite{splitsr2021imwut}, optimized for different targets: \textbf{SplitSR-accuracy} and \textbf{-latency} (Appendix \ref{app:splitsr}).
% 
% \begin{itemize}[leftmargin=*,noitemsep,topsep=0pt]
%     \item \textbf{SplitSR-accuracy}: The accuracy-optimized model, composed of 7 residual groups, each with 7 residual blocks.
%     \item \textbf{SplitSR-latency}: The latency-optimized model, composed of 5 residual groups, each with 6 residual blocks.
% \end{itemize}
SplitSR introduces a compact residual block (SplitSRBlock) and modifies RCAN to allow for a configurable accuracy-computational cost trade-off, using a single model. Two system configurations were presented in \cite{splitsr2021imwut}, optimized for different targets:

\begin{itemize}[leftmargin=*,noitemsep,topsep=0pt]
    \item \textbf{SplitSR-accuracy}: The accuracy-optimized model, composed of 7 residual groups, each with 7 residual blocks.
    \item \textbf{SplitSR-latency}: The latency-optimized model, composed of 5 residual groups, each with 6 residual blocks.
\end{itemize}

Moreover, SplitSR is optimized for mobile CPU execution through the TVM compiler~\cite{tvm_2018odsi}. To compare against SplitSR, we impose a PSNR constraint within 0.05 dB of the PSNR achieved by each SplitSR variant and select the \tool model that satisfies it for each dataset. As such, we select IMDN and MobiSR-RCAN to compare with SplitSR-accuracy and -latency, respectively (Table~\ref{tab:eval_sota}). 

\begin{figure}[t]
    \centering
    % \vspace{-0.5cm}
    \subfloat[Comparison on B100.]{
        \centering
        \includegraphics[trim={0.0cm 0.8cm 1cm 1cm},clip,width=0.235\textwidth]{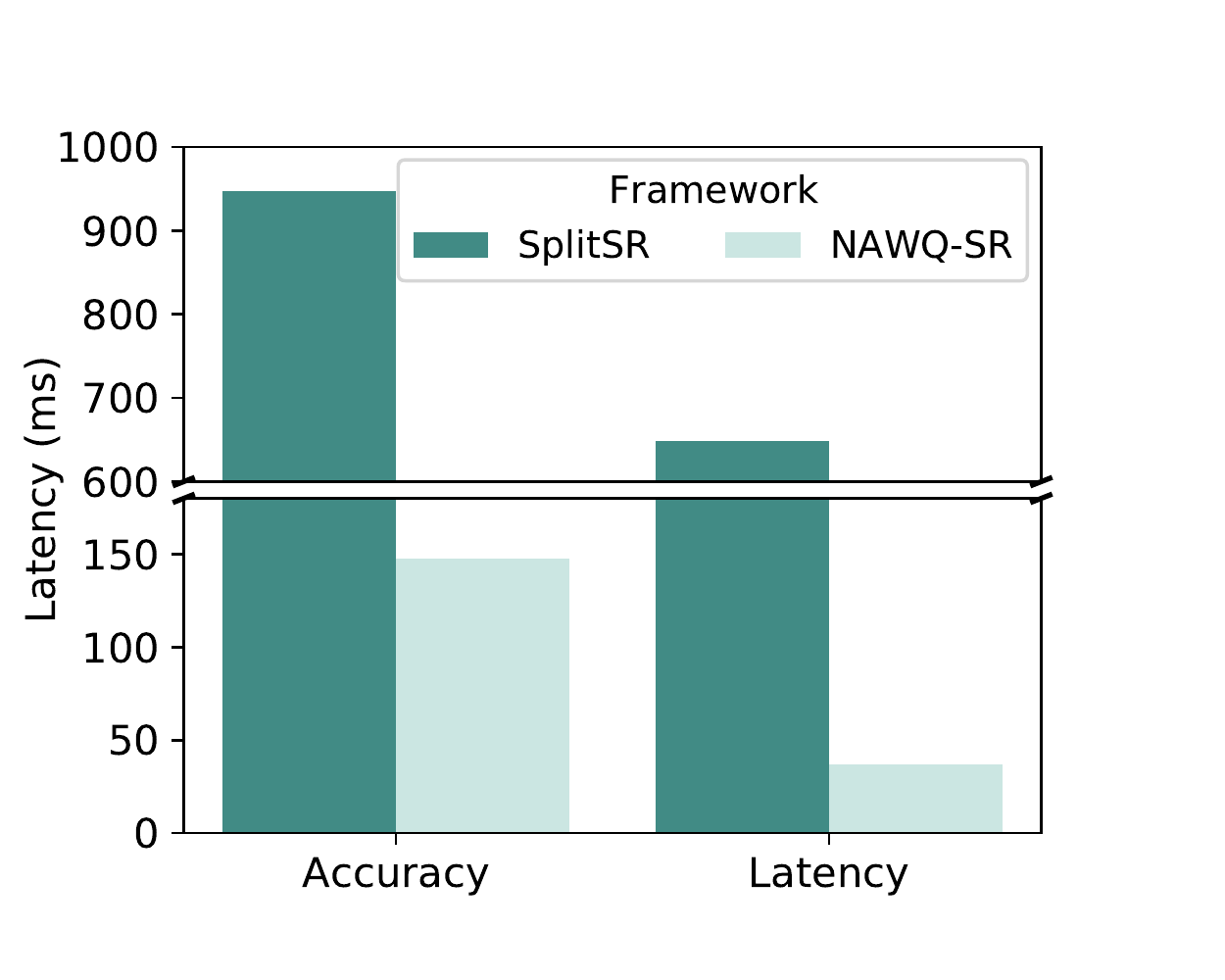}
        % \vspace{-0.7cm}
        \label{fig:splitsr_b100}
    }
    \subfloat[Comparison on Urban100.]{
        \centering
        \includegraphics[trim={0.0cm 0.8cm 1cm 1.0cm},clip,width=0.235\textwidth]{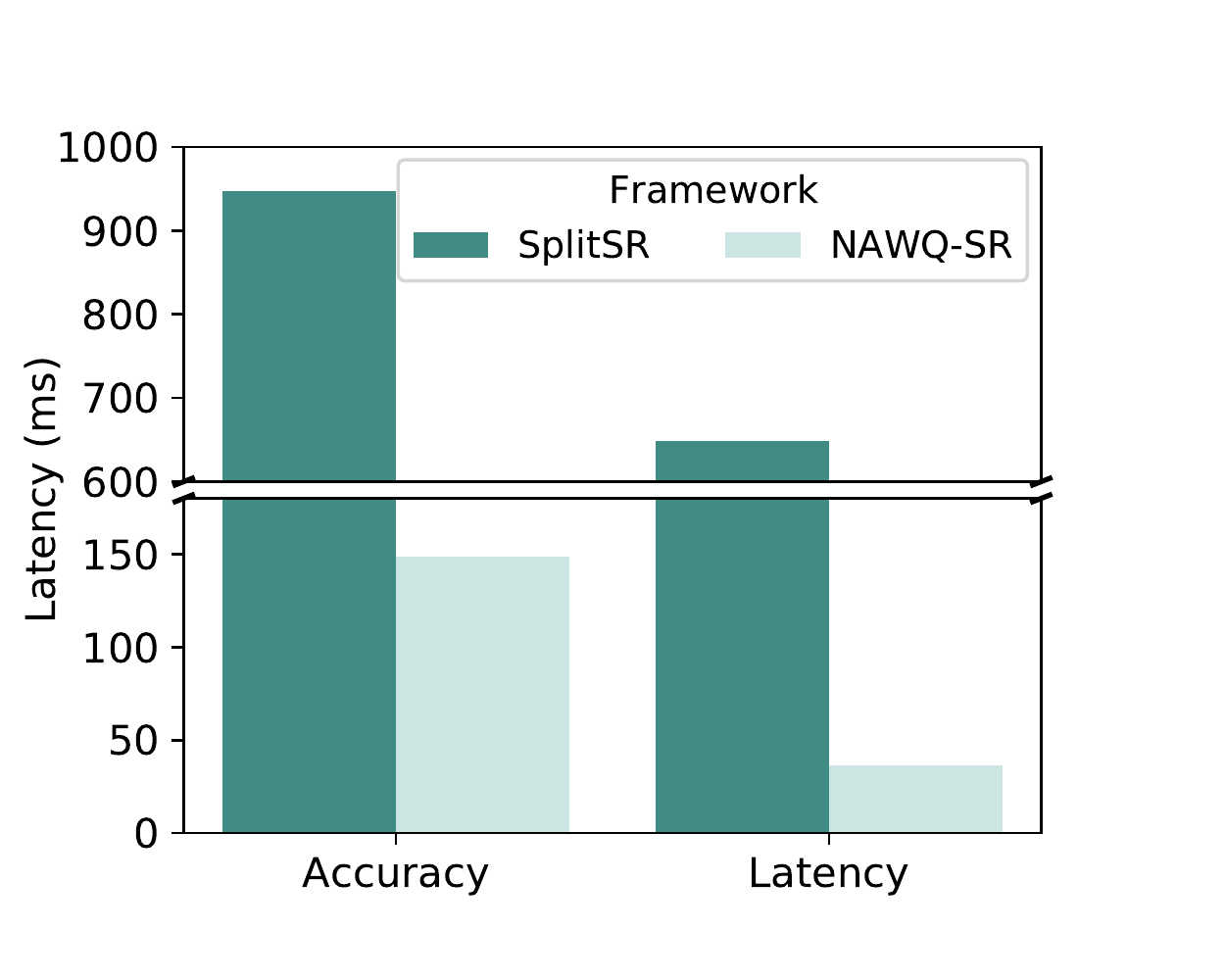}
        % \vspace{-0.7cm}
        \label{fig:splitsr_urban100}
    }
    \vspace{-0.2cm}
    \caption{Latency comparison against SplitSR.}
    \vspace{0.1cm}
    \label{fig:splitsr}
\end{figure}

Fig.~\ref{fig:splitsr} shows the measured latency of SplitSR and \tool on Urban100 and B100. On the accuracy-driven designs, \tool improves latency by 1.59$\times$ and 1.60$\times$ on Urban100 and B100, respectively. On latency-driven designs, \tool demonstrates a performance gain of 4.40$\times$ and 4.37$\times$ over SplitSR on Urban100 and B100, respectively. As a result, although SplitSR effectively combines a lightweight model design together with compiler optimizations to achieve significant speedup, it still relies on CPU execution, remaining bounded by the performance of floating-point processors. 
On the other hand, \tool's hybrid precision and optimized utilization of the NPU's processing units avoids the inefficiencies of floating-point execution and reaches higher raw performance over the highly optimized CPU-based SplitSR.

% \vspace{-0.2cm}
\textbf{Comparison with XLSR.} For fairness, we target the similarly sized TPSR with the same upscaling factor ($\times$3) as XLSR (Table~\ref{tab:eval_sota}-bottom). \tool outperforms the INT8 XLSR with 91\% higher speedup. \blue{This can be attributed to the fact that XLSR changes the number of channels of the convolutional layers quite frequently along the DNN model as a way of balancing computational cost and model capacity. Despite the theoretical reduction in FLOP count, this has been shown to lead to increased cache-miss rates
~\cite{shufflenetv2_2018eccv} and in turn to increased latency on existing NPUs. Instead, \tool allows existing models to run without architectural modifications by providing latency gains through its hybrid-precision execution. As such, it does not require from DNNs to frequently change the number of channels across convolutional layers, leading to more efficient NPU execution.}

With respect to quality, XLSR reports a drop between 0.2-0.5~dB when quantizing to INT8~\cite{xlsr2021cvprw}. \tool achieves significant PSNR gains of \mbox{0.21-0.51~dB} and 0.27-0.57~dB over the INT8 XLSR on B100 and Urban100, respectively, while yielding same or higher PSNR levels over the FP32 XLSR.
This can be attributed to the fact that XLSR replaces pixel-shuffle blocks with transpose convolutions in order to avoid the lack of support for pixel-shuffling on NPUs. In turn, this leads to checkerboard artifacts and hence deteriorates the achieved visual quality~\cite{deconv_checkerboard_artifacts}. On the other hand, \tool's memory-aware mapping for efficiently executing pixel-shuffle blocks on the NPU (\S~\ref{sec:neural_codec}) leads to both lower latency and higher visual quality, setting a new state-of-the-art in latency-quality for NPU-based SR.
% \manote{missing the insight why}

% \vspace{-0.4cm}
\vspace{-0.5em}
\subsection{Energy Consumption}
\label{sec:eval_energy}
% \vspace{-0.1cm}

% Next, we compare the energy consumption of \tool's NPU-optimized hybrid-precision execution against the status-quo of CPU/GPU/NPU execution.
To evaluate \tool's energy efficiency, we processed 50 images using TPSR and MobiSR-RCAN.
% , for each of the execution schemes.
The images are pre-hosted, representing the scenario where a user would have a downloaded content, which is then enhanced with on-device SR. 
% and visualized in high resolution. 
Energy was measured with the Monsoon power monitor~\cite{monsoon} at a sampling period of 200 $\mu$s.

Fig.~\ref{fig:energy} shows the average energy consumption for the two models when upscaling to 720p images \red{on SDM865 and SDM888}.
In this case, we subtract the average idle energy when the screen is on.
We observe that \tool results in significant energy savings compared to the FP32 CPU execution, with an average $6.1\times$ and $10.3\times$ reduction per model on SDM865 and \red{8.5$\times$ and 14.2$\times$ on SDM888}.
This result motivates the adoption of NPU-optimized frameworks in comparison to state-of-the-art CPU-centric on-device SR approaches, such as SplitSR.
Moreover, we see a significant $3.5\times$-$4.3\times$ and \mbox{$2.1\times$-$2.4\times$} energy reduction, even when compared to the more efficient FP16 GPU and A16W8 NPU, respectively, \red{with similar gains observed for SDM888}.

Fig.~\ref{fig:battery} estimates the battery life when a user continuously watches SR-enhanced video at 1080p on a device with $4000$mAh, a common battery capacity for recent mobile devices (\textit{e.g.}~Samsung S20). In this case, we measure the total energy, including the screen consumption.
\tool greatly prolongs the battery life, with up to $3.8\times$, $2.3\times$ and $1.9\times$ battery life extension when compared to CPU, GPU and A16W8 NPU execution, respectively. \red{When targeting SDM888, we observe similar gains, with a slight improvement due to the larger hardware improvement of SDM888's NPU perfomance over the CPU.}
This result highlights the potential for existing state-of-the-art end-to-end on-device SR systems, such as NEMO, which are bounded to GPU-based execution due to visual quality constraints, to integrate \tool as a means of improving not only latency and visual quality, but also extending battery life.

% \begin{figure}[t]
%     \captionsetup[subfigure]{labelformat=empty}
%     \centering
%     \vspace{-0.3cm}
%     \subfloat[]{
%         \includegraphics[
%         width=0.4\textwidth]{figures/quality-anal/IMDN-Urban100-45.pdf}
%     }\\
%     % \vspace{-1cm}
%     \vspace{-0.6cm}
%     \centering
%     \subfloat[]{
%         \includegraphics[
%         width=0.4\textwidth]{figures/quality-anal/RCAN-Urban100-23.pdf}
%     }\\
%     \vspace{-0.7cm}
%     \caption{Qualitative comparison between INT8 and \tool $\times4$ upscaling. INT8 results in visual artifacts on both textures and colors when compared to \tool's hybrid-precision and DRE-based quality enhancement.
%     }
%     \vspace{0.1cm}
%     \label{fig:qualitative}
% \end{figure}

\section{Discussion}
\label{sec:discussion}

\textbf{\tool and existing mixed-precision schemes.} 
Recently, the ML community has studied a range of mixed-precision quantization schemes that, similarly to \tool, assign a different bitwidth to each layer. Focusing on the strategy of selecting the layerwise bitwidth and following the taxonomy of Huang~\textit{et al.}~\cite{sqd2022icml}, we discuss \textit{i)}~search-based, \textit{ii)}~metric-based, and \textit{iii)}~optimization-based methods. 

Search-based methods typically rely on neural architecture search (NAS) or reinforcement learning (RL) algorithms in order to yield the layerwise bitwidths. As noted in \S~\ref{sec:background} with the example of HAQ~\cite{wang2019haq}, this family of techniques introduces a significant computational overhead and requires re-training, making it unsuitable for post-training deployment of pre-trained SR models.

Metric-based methods assign bitwidths by estimating the layerwise resilience to low precision with metrics that are relatively cheap to calculate, such as the Hessian-based metric adopted by HAWQ~\cite{hawq2019iccv,hawqv22020neurips}. Despite the reduced computational burden, existing metric-based methods still require either quantization-aware training or a re-training stage and hence cannot be applied post-training to existing models.
Finally, the optimization-based methods aim to turn the wordlength selection to a differentiable optimization problem~\cite{fracbits2021aaai, sqd2022icml}. This line of work can lead to a more stable quantization-aware training process when using mixed-precision. However, these methods have been tailored only for cases where quantization-aware training can be performed.

Despite the various merits of these methods, their effectiveness relies on \textit{training-time} techniques or requires a \textit{re-training step}. For both of these to be feasible, availability of the training set is required. With privacy concerns increasing by both users and service providers, this assumption is often not valid, as in the case of strict privacy regulations for sharing user data~\cite{gdpr2018}, privacy-centric initiatives by service providers~\cite{apple} or confidentiality clauses over proprietary datasets collected by industrial companies.

In this context, \tool introduces a wordlength selection method that requires a minimal calibration set and enables the use of hybrid precision in cases where the training set is \textit{not} available. As such, our work offers the computational efficiency of metric-based techniques, but can also be applied directly on pre-trained models \textit{post-training}.

\begin{figure}[t]
    \centering
    % \vspace{-0.5cm}
    % \vspace{-0.9cm}
    \subfloat[Energy usage on SDM865.]{
        \centering
        \includegraphics[trim={0cm 0.8cm 1cm 1cm},clip,width=0.22\textwidth]{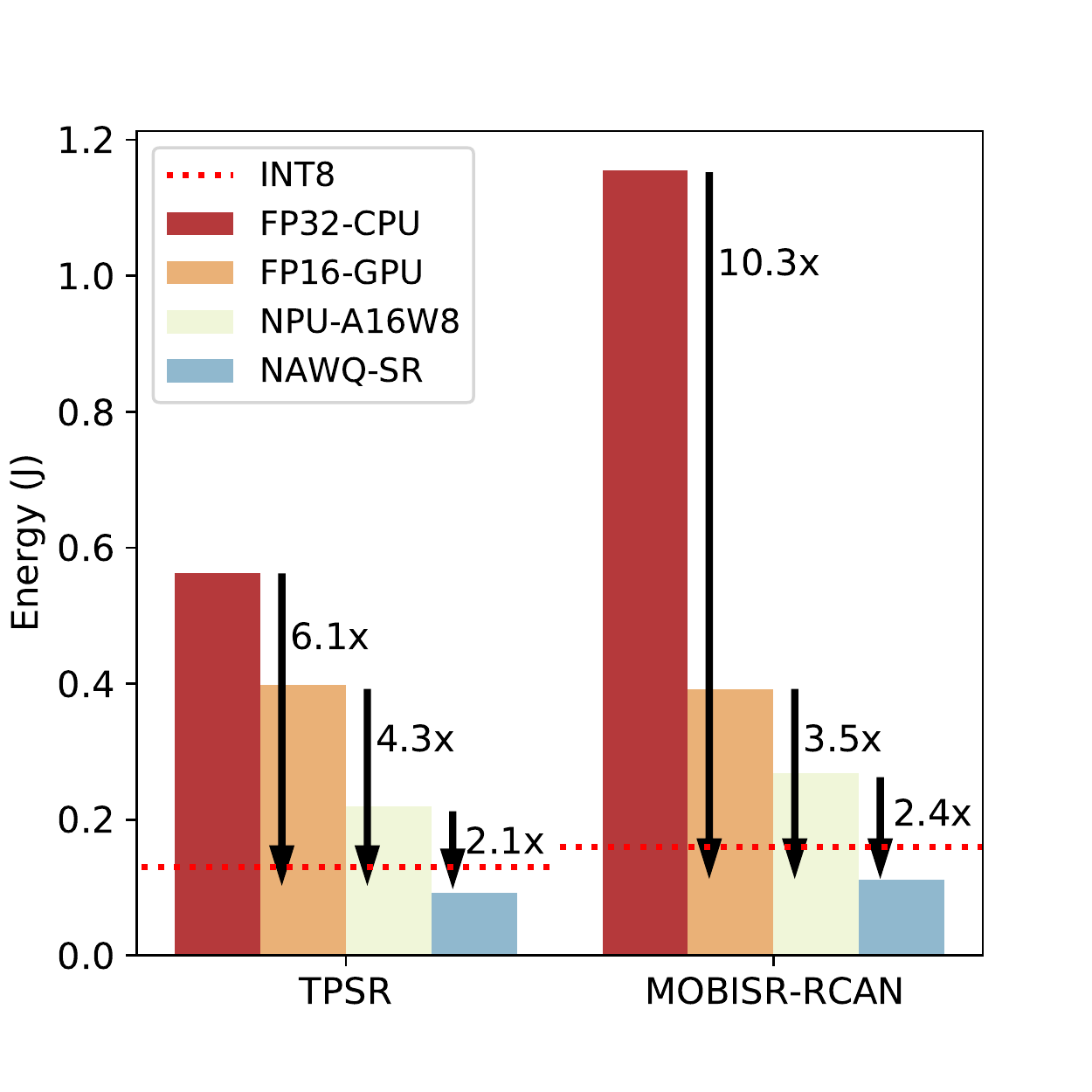}
        \label{fig:energy}
    }
    \subfloat[Battery life on SDM865.]{
        \centering
        \includegraphics[trim={0.0cm 0.8cm 1cm 1.0cm},clip,width=0.22\textwidth]{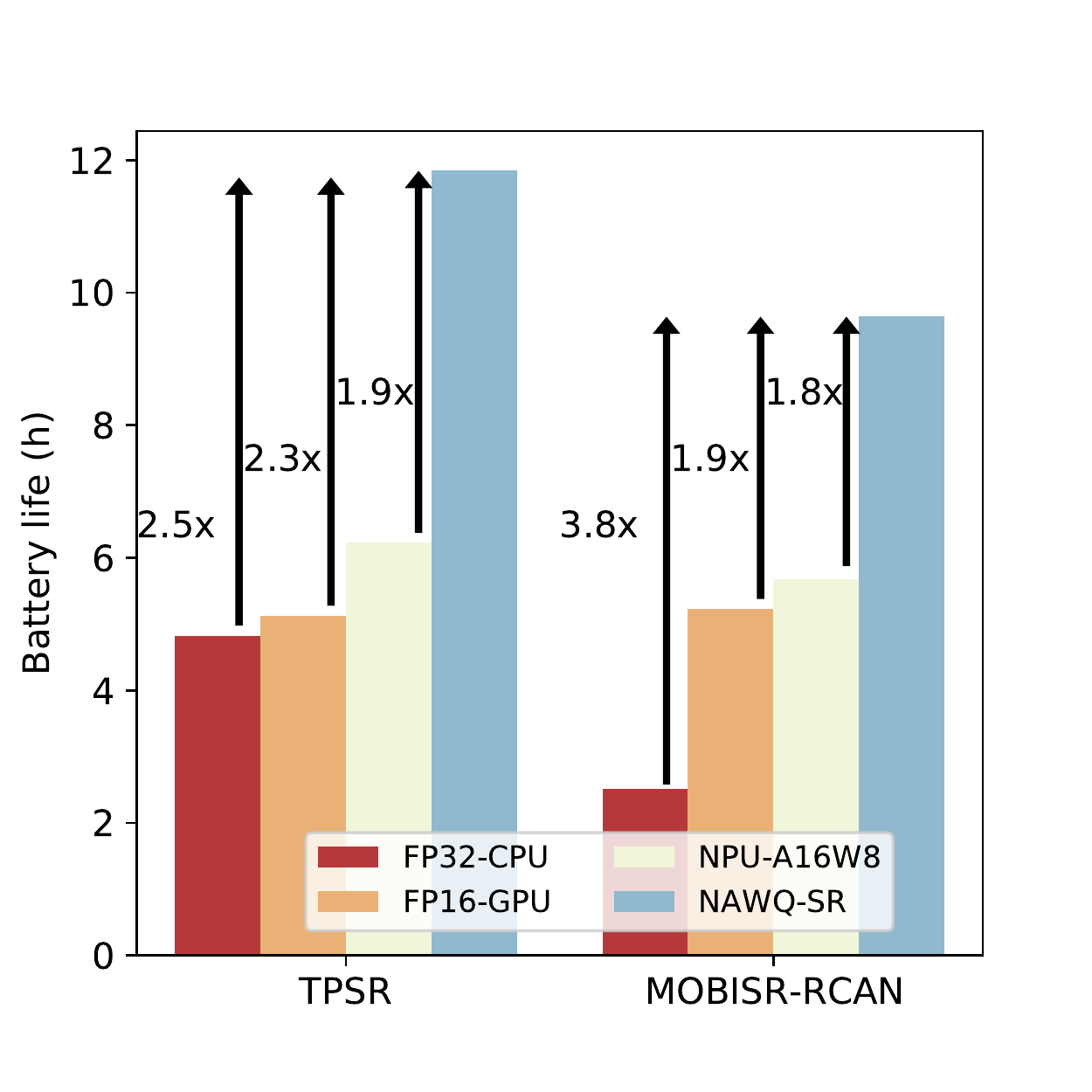}
        \label{fig:battery}
    }

    \subfloat[\red{Energy usage on SDM888.}]{
        \centering
        \includegraphics[trim={0cm 0.8cm 1cm 1cm},clip,width=0.22\textwidth]{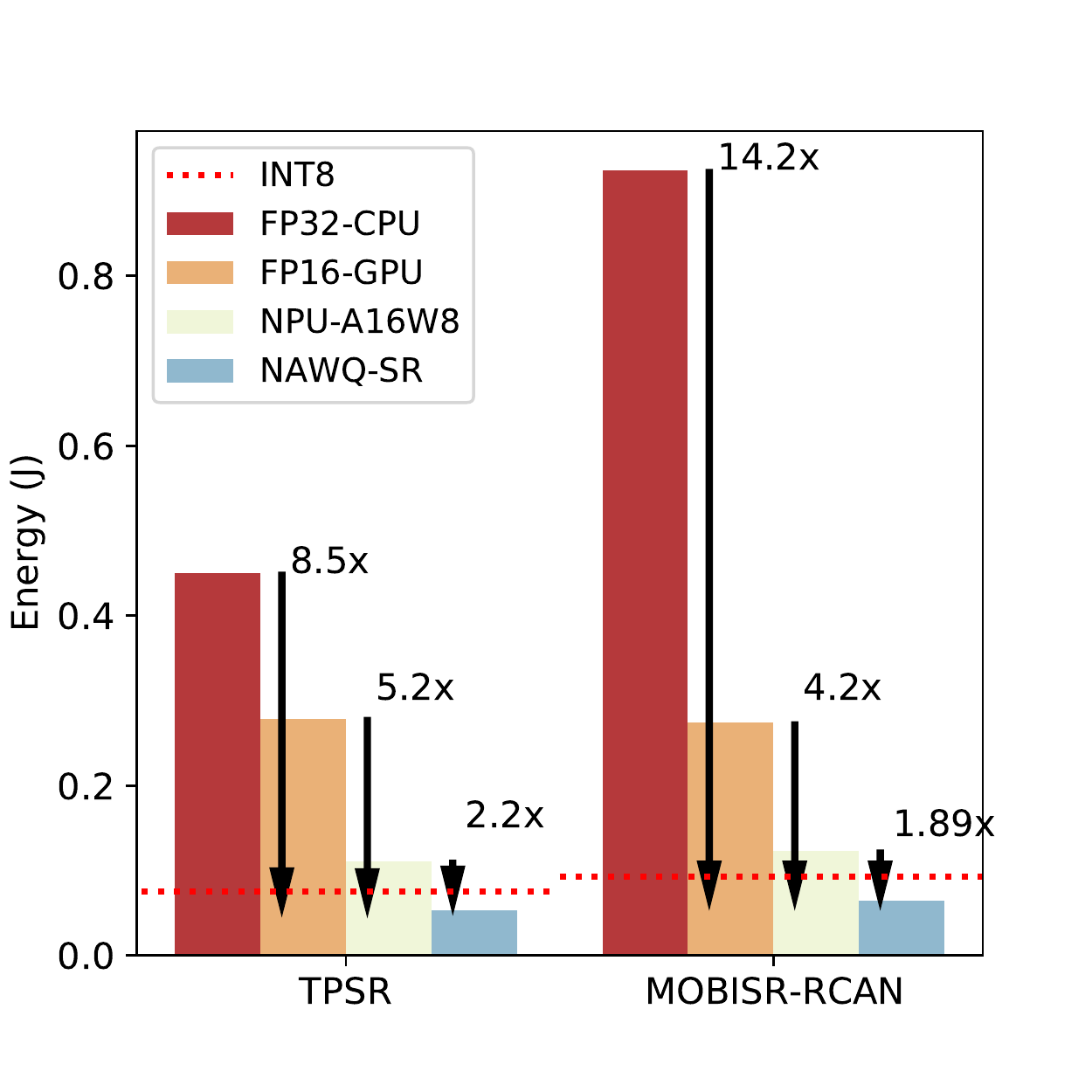}
        \label{fig:energy}
    }
    \subfloat[\red{Battery life on SDM888.}]{
        \centering
        \includegraphics[trim={0.0cm 0.8cm 1cm 1.0cm},clip,width=0.22\textwidth]{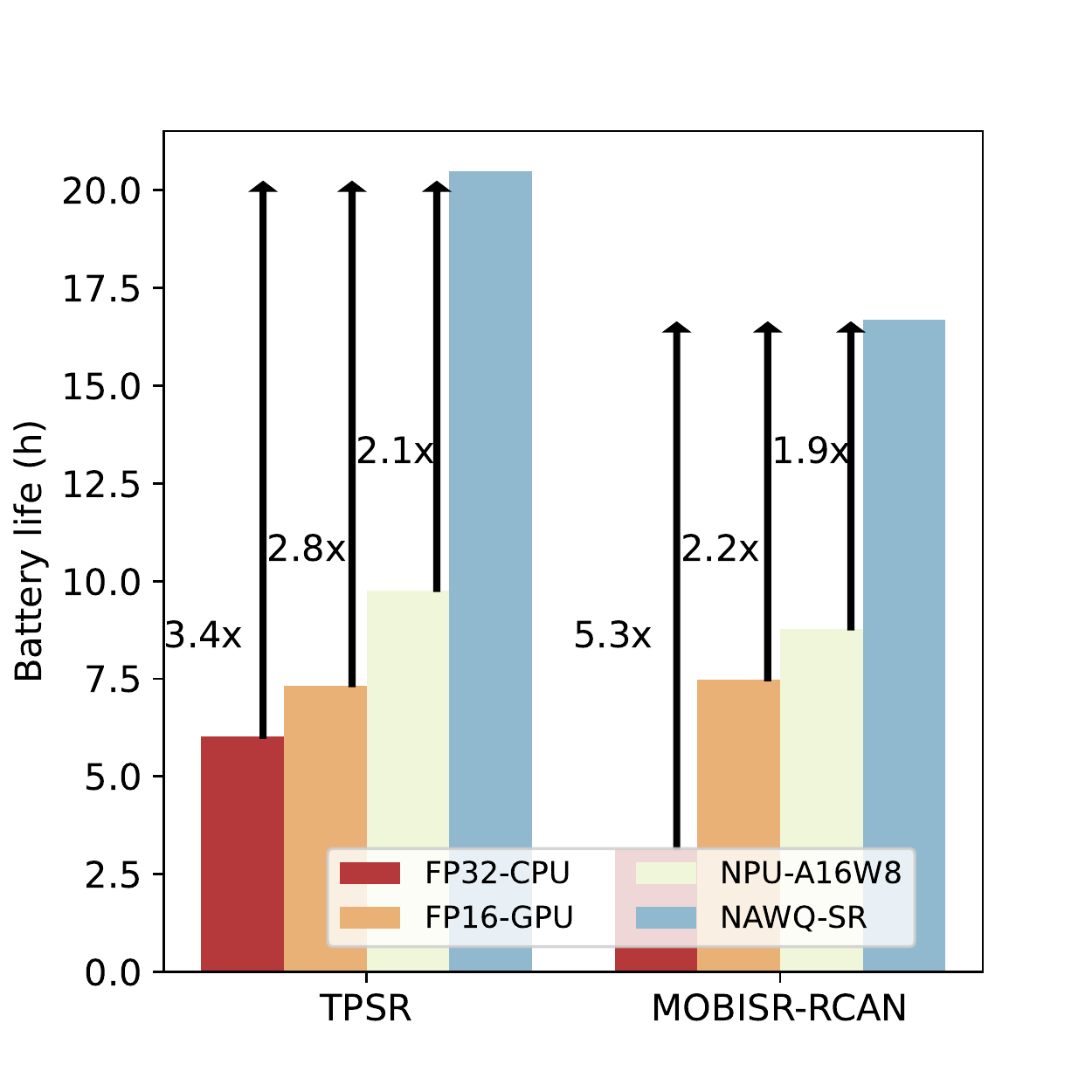}
        \label{fig:battery}
    }
    
    \vspace{-0.2cm}
    \caption{Energy consumption and battery life comparison for 720p and 1080p content, respectively.}
    \vspace{0.1cm}
    \label{fig:energy_battery}
\end{figure}

\textbf{Applicability to other mobile NPUs.} In this work, we \red{primarily} targeted the NPU of Qualcomm's SDM865 SoC as a representative mobile NPU with available software tools. SDM865 has a vendor-specific hardware architecture, comprising two distinct units, HVX and HTA (see \S~\ref{sec:eval}), that support INT8 and both INT8 and A16W8 execution, respectively. As such, if \tool was relying on the existence of two distinct units to obtain its performance, it would have narrow applicability to dual-unit NPUs.

On the contrary, \tool does not require the existence of two distinct units. \tool's processing flow and neural image codec is designed for NPU hardware architectures with either one and two processing units. \red{In our evaluation, we demonstrate this generality of our framework by targeting also the NPU of SDM888, which comprises a single composable processing unit.}
In a similar fashion, a broad range of existing mobile NPUs, such as the Samsung Exynos NPU~\cite{samsung_npu2021isca}, MediaTek APU~\cite{mediatek_apu2020isscc} and Arm Ethos~\cite{arm_ethos_npu_perf}, consist of a \textit{single} processing unit that can be configured with either INT8 or A16W8 at run time. Hence, by not introducing any optimizations that are coupled to two processing units executing with different wordlength, \tool constitutes an on-device SR framework that is generalizable across mobile NPUs from different vendors.

\red{Despite the general underlying principles of our framework, mobile NPUs are often characterized by heterogeneity in terms of both hardware and software~\cite{ai_benchmark2019iccvw, facebook2019hpca}. As such, it is difficult to deploy our method out of the box without any further engineering step; to obtain the gains demonstrated by \tool, its runtime components may have to be adapted and optimized based on the available API of the target NPU. Nonetheless, with interoperability across diverse mobile SoCs being an active area of research~\cite{oodin2021smartcomp}, it constitutes an important, yet orthogonal, consideration when attempting to deploy our framework on new NPU-equipped SoC architectures.}

\vspace{-0.1cm}
\section{Conclusion}
\label{sec:conclusion}
% \vspace{-0.2cm}

% The \tool framework introduces both algorithmic and system optimization techniques to achieve state-of-the-art SR on mobile NPUs.
% Our experiments show that our proposed hybrid-precision %wordlength optimization 
% method can efficiently scale to SR models of varying computational complexity, making \tool applicable to any given SR model. %enabling \tool to be applicable to any given SR model.
% We also highlight that our run-time adaptive precision technique can be effectively deployed in existing commercial NPUs by means of \tool's neural image codec, resulting in quality gains with minimal overhead.

% As a stand-alone framework, \tool surpasses the performance of existing on-device SR systems, overcoming their limitations and mitigating the quality drawbacks of executing SR DNNs on low-precision units.
% Additionally, \tool can be orthogonally combined with existing frameworks to obtain further gains, by either enabling them to target NPUs, \textit{e.g.}~for the CPU-based SplitSR and GPU-based NEMO, or with better utilization of the NPU resources, \textit{e.g.}~for MobiSR's NPU-mapped compact model. 

%%% CHECK. Otherwise uncomment old conclusion above
% \tool introduces both algorithmic and system optimization techniques on mobile NPUs, surpassing the performance of existing on-device SR systems by mitigating the quality drawbacks of executing SR DNNs on low-precision units.
\tool introduces both algorithmic and system optimization techniques on mobile NPUs in order to mitigate the quality drawbacks of executing SR DNNs on low-precision units. 
Our experiments show that our proposed hybrid-precision method can scale to SR models of varying computational complexity and the run-time precision adaptation method of \tool's neural image codec can be efficiently deployed in existing commercial NPUs.

As a stand-alone framework, \tool surpasses the performance of existing on-device SR systems, overcoming their limitations and significantly mitigating the quality drawbacks of executing SR DNNs on low-precision units.
% Apart from surpassing the performance of existing on-device SR systems, \tool can be orthogonally combined with these systems to enable them to better target NPUs for further gains.
Additionally, \tool can be orthogonally combined with existing frameworks to obtain further gains, by either enabling them to target NPUs, \textit{e.g.}~for the CPU-based SplitSR and GPU-based NEMO, or with better utilization of the NPU resources, \textit{e.g.}~for MobiSR's NPU-mapped compact model.

\ifCLASSOPTIONcompsoc
  \section*{Acknowledgments}
  This work was supported by Samsung AI and the European Research Council via the REDIAL project.
\else
  \section*{Acknowledgment}
\fi


% % Can use something like this to put references on a page
% % by themselves when using endfloat and the captionsoff option.
% \ifCLASSOPTIONcaptionsoff
%   \newpage
% \fi

% % trigger a \newpage just before the given reference
% % number - used to balance the columns on the last page
% % adjust value as needed - may need to be readjusted if
% % the document is modified later
% %\IEEEtriggeratref{8}
% % The "triggered" command can be changed if desired:
% %\IEEEtriggercmd{\enlargethispage{-5in}}

% % references section

% % can use a bibliography generated by BibTeX as a .bbl file
% % BibTeX documentation can be easily obtained at:
% % http://mirror.ctan.org/biblio/bibtex/contrib/doc/
% % The IEEEtran BibTeX style support page is at:
% % http://www.michaelshell.org/tex/ieeetran/bibtex/
% %\bibliographystyle{IEEEtran}
% % argument is your BibTeX string definitions and bibliography database(s)
% %\bibliography{IEEEabrv,../bib/paper}
% %
% % <OR> manually copy in the resultant .bbl file
% % set second argument of \begin to the number of references
% % (used to reserve space for the reference number labels box)


% \end{thebibliography}

\bibliographystyle{IEEEtran}
\bibliography{references}

\end{document}